%% file: iclr2027_conference.tex
\documentclass{article} 

\usepackage{iclr2027_conference,times}

\input{math_commands.tex}

\usepackage{hyperref}
\usepackage{url}

\title{Can Learning to Answer Spatial Questions\\Improve Novel-View Generation?}

\author{
Jahyeok Koo \quad
Hyeonseo Yu \quad
Seungho Jang \quad
Jaewoo Jung \quad 
Jiwon Kang \\  
\ \bf Seokju Cho \quad Seungryong Kim\textsuperscript{\(\dagger\)}\\[2mm]
\ KAIST AI\\[1mm]
\ Project page: \url{https://cvlab-kaist.github.io/Ask4NV/}%
}

\input{preamble}
\iclrfinalcopy 
\begin{document}

\maketitle

\vspace{-25pt}
\begin{figure}[h]
  \centering
  \includegraphics[width=\linewidth]{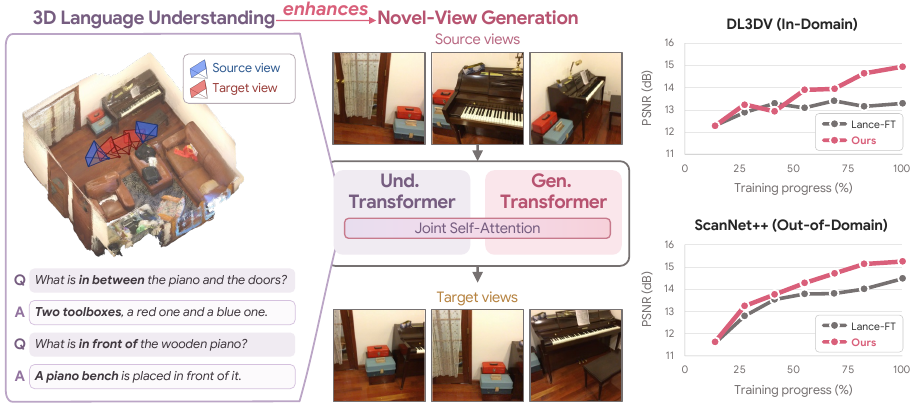}
  \vspace{-10pt}
  \caption{We present \textbf{\ours}, which enhances novel-view generation through 3D language understanding. Within a Mixture-of-Transformers architecture, 3D language understanding supervision trains the understanding transformer to interpret scene semantics and spatial relationships. During novel-view generation, joint self-attention combines this knowledge with the generation transformer's visual synthesis capabilities.
  \ours achieves higher PSNR than the baseline trained solely on novel-view generation on both in-domain and out-of-domain datasets.}
  \label{fig:teaser}
\end{figure}

\begingroup
\renewcommand{\thefootnote}{\ensuremath{\dagger}}
\footnotetext{Corresponding author.}
\endgroup

\input{sec/0_abstract}
\input{sec/1_intro_nvs_}
\input{sec/3_Preliminaries}
\input{sec/4_methods}

\input{sec/5_experiments}
\input{sec/6_conclusions}
\clearpage
\bibliography{iclr2027_conference}
\bibliographystyle{iclr2027_conference}

\appendix
\input{sec/X_appendix}

\end{document}

%% file: math_commands.tex
\usepackage{amsmath,amsfonts,bm}

\def\eqref#1{equation~\ref{#1}}

\def\1{\bm{1}}

\DeclareMathAlphabet{\mathsfit}{\encodingdefault}{\sfdefault}{m}{sl}
\SetMathAlphabet{\mathsfit}{bold}{\encodingdefault}{\sfdefault}{bx}{n}



%% file: sec/0_abstract.tex
\begin{abstract}
Generating novel views from limited observations requires preserving both the appearance and spatial structure of a scene.
Although recent approaches exploit fine-grained geometric cues, they do not explicitly model high-level semantic and spatial properties such as object identity, attributes, and inter-object relationships.
In this work, we investigate whether 3D language understanding can provide complementary supervision for novel-view generation.
We introduce \textbf{\ours}, built on a pretrained unified multimodal model with a Mixture-of-Transformers (MoT) architecture.
Our model incorporates camera geometry into both VAE and ViT representations and leverages understanding features during generation through joint self-attention with a modality-aware attention masking strategy.
Experiments demonstrate strong image quality and geometric consistency in novel-view generation, alongside competitive 3D language understanding.
Further analysis shows that joint training improves cross-view correspondence in VAE representations and similarity in a 3D-aware feature space.

\end{abstract}

%% file: sec/1_intro_nvs_.tex
\etocdepthtag.toc{main}
\section{Introduction}

Novel-view generation~\citep{gao2024cat3d, lu2025matrix3d, kwon2025cameo} uses generative priors learned from large-scale data to synthesize images of a scene from unseen viewpoints given limited observations, while preserving scene appearance and geometric consistency.
Recent approaches incorporate geometric information through depth-based warping, cross-view correspondence or explicit geometry supervision~\citep{seo2024genwarp, kwon2025cameo, lu2025matrix3d}, providing fine-grained cues for cross-view consistency.
These advances support a wide range of applications, including video editing~\citep{zhang2025recapture}, robot manipulation~\citep{kim2025dreamgrasp}, and vision-language driving~\citep{wang2025dreamnav}.

While such fine-grained geometric cues reveal the scene structure and dense correspondences across viewpoints, they do not explicitly offer the interpretation of high-level scene properties, such as object identities, attributes, and inter-object relationships.
Such high-level understanding can also guide novel-view generation by informing how objects and their spatial arrangements should appear from a different viewpoint.
For example, recognizing a red chair beside a wooden table can guide the model to preserve the chair's identity, attributes, and relationship to the table, even as its visible shape and image position change with the viewpoint.
Learning these semantic and spatial properties could therefore complement fine-grained geometric information, helping the model preserve scene content and represent its spatial organization consistently across views.

However, the potential of explicitly learning such high-level properties for novel-view generation remains largely unexplored.
We then ask: \textit{How can we provide such high-level semantic and spatial supervision, and how can it be incorporated into joint learning with novel-view generation?}
We investigate this direction through 3D language understanding tasks~\citep{hong20233d, huang2023embodied, zheng2025video}, which involve identifying and localizing objects and reasoning about their attributes and spatial relationships.
Fig.~\ref{fig:concept} highlights the shared semantic and spatial properties underlying 3D language understanding and novel-view generation, suggesting the potential of language-based supervision to support generation.
To jointly learn novel-view generation and 3D language understanding, we adopt pretrained unified multimodal models~\citep{xie2025show,team2024chameleon}, which support both visual generation and multimodal understanding within a single model.

\input{figures/tex/concept}

Concretely, we introduce \textbf{\ours}, which builds on a pretrained Mixture-of-Transformers (MoT) architecture~\citep{liang2024mixture, deng2025emerging} and introduces a small set of architectural modifications to jointly model 3D language understanding and novel-view generation.
The MoT architecture provides task-specific parameters together with joint self-attention between the two transformers, allowing the two tasks to be learned within a shared framework without introducing a separate cross-task fusion module.
For novel-view generation, we use VAE latents alongside complementary ViT features from the understanding transformer and incorporate camera geometry into both representations through Pl\"ucker ray embeddings.
We further design an attention masking strategy to facilitate information exchange across different token modalities during novel-view generation.

Our analyses examine how joint training with 3D language understanding affects the model's visual representations and novel-view generation capabilities.
Extensive experiments show that our model achieves strong novel-view generation performance in both image quality and geometric consistency across multiple evaluation settings.
It also performs competitively on 3D language understanding despite the limited performance of its pretrained backbone on these tasks.
Ablation studies further validate our architectural and attention design choices.

%% file: figures/tex/concept.tex
\begin{figure*}[t]
    \centering
    \includegraphics[width=\linewidth]{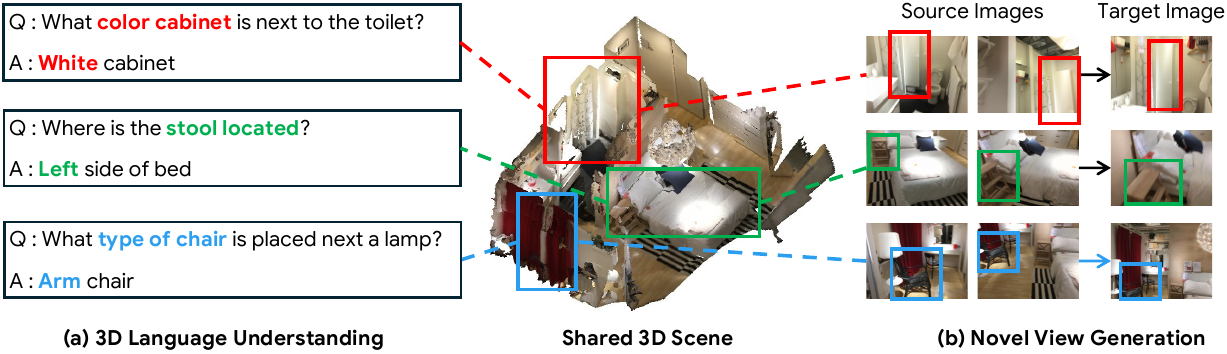}
    \caption{
        \textbf{Shared scene properties across 3D language understanding and novel-view generation.} Questions about object categories, attributes, and spatial relationships address the same scene properties that should be represented in novel views. The knowledge required to answer these questions can provide complementary semantic and spatial supervision for novel-view generation.
    }
    \label{fig:concept}
    \vspace{-10pt}
\end{figure*}

%% file: sec/3_Preliminaries.tex
\section{Preliminaries}

\paragrapht{Novel-view generation.}
We follow a common formulation of prior novel-view generation approaches~\citep{gao2024cat3d, kwon2025cameo}, which generate target images given a set of source images and the camera parameters of both source and target views.
Leveraging generative priors, these approaches aim to synthesize unseen views that preserve scene appearance and maintain geometric consistency, including interpolated and challenging extrapolated viewpoints.
Formally, let $\mathbf{I}_{\mathrm{src}}$ denote a set of $N$ source images and $\mathbf{C}_{\mathrm{src}}$ their corresponding camera parameters.
Given the camera parameters $\mathbf{C}_{\mathrm{tgt}}$ for $M$ target views, the model generates the corresponding target images:
\begin{equation}
\hat{\mathbf{I}}_{\mathrm{tgt}}
=
\mathcal{G}_{\theta}
\left(
\mathbf{I}_{\mathrm{src}},
\mathbf{C}_{\mathrm{src}},
\mathbf{C}_{\mathrm{tgt}}
\right),
\end{equation}
where $\mathcal{G}_{\theta}$ denotes the novel-view generation model.

In this work, we formulate novel-view generation using flow matching~\citep{lipman2022flow}.
Specifically, a VAE encoder maps the source images $\mathbf{I}_{\mathrm{src}}$ into latent representations $\mathbf{z}_{\mathrm{src}}=\mathcal{E}_{\mathrm{vae}}(\mathbf{I}_{\mathrm{src}})$.
Given noisy target latents $\mathbf{z}_t$ at timestep $t$, the model predicts the corresponding velocity field conditioned on the source latents and the camera parameters of the source and target views.
The model is trained by minimizing the squared error between the predicted velocity $\hat{\mathbf{v}}_t$ and the target velocity $\mathbf{v}_t^{*}$:
\begin{equation}
    \hat{\mathbf{v}}_t
    =
    \mathbf{v}_{\theta}
    \left(
        \mathbf{z}_t, t;
        \mathbf{z}_{\mathrm{src}},
        \mathbf{C}_{\mathrm{src}},
        \mathbf{C}_{\mathrm{tgt}}
    \right),
    \qquad
    \mathcal{L}_{\mathrm{gen}}
    =
    \mathbb{E}
    \left[
        \left\|
        \hat{\mathbf{v}}_t - \mathbf{v}^{*}_t
        \right\|_2^2
    \right],
\end{equation}
where $\mathbf{v}_{\theta}$ denotes the velocity prediction function, and the expectation is taken over training samples, timesteps, and noise.
At inference, target latents are initialized from Gaussian noise, updated by integrating the predicted velocity field, and decoded into images using the VAE decoder.

\paragrapht{3D language understanding.}
We consider 3D question answering~\citep{ma2022sqa3d, azuma2022scanqa} and visual grounding~\citep{zhang2023multi3drefer, chen2020scanrefer} as two representative 3D language understanding tasks for reasoning about scene semantics and spatial relationships through language.
Given multi-view images and their associated 3D coordinates, the model generates a textual answer to a scene-related question or identifies and localizes objects referred to by a language description.

Specifically, the input images are encoded by a ViT encoder and mapped to the language model's hidden space through a visual projector,
$\mathbf{e}^{\mathrm{vit}}=P_{\mathrm{vit}}(\mathcal{E}_{\mathrm{vit}}(\mathbf{I}))$.
The associated 3D coordinates are incorporated through positional embeddings to obtain spatially enriched visual tokens
$\tilde{\mathbf{e}}^{\mathrm{vit}}$.
The input question is tokenized and mapped to the same hidden space through the language model's embedding layer, yielding
$\mathbf{e}^{\mathrm{text}}$.
Conditioned on these visual and text embeddings, the model generates an answer $\mathbf{y}=(y_1,\ldots,y_T)$ autoregressively and is trained with the cross-entropy loss over the ground-truth answer tokens:
\begin{equation}
    \mathcal{L}_{\mathrm{qa}}
    =
    -\sum_{i=1}^{T}
    \log p_{\theta}
    \left(
        y_i
        \mid
        y_{<i},
        \tilde{\mathbf{e}}^{\mathrm{vit}},
        \mathbf{e}^{\mathrm{text}}
    \right).
\end{equation}

For visual grounding, training is performed with a proposal-matching objective $\mathcal{L}_{\mathrm{grd}}$, whose formulation and implementation details are provided in the supplementary material (Sec.~\ref{supp:3d_und}).

\paragrapht{Shared scene properties across tasks.}
Fig.~\ref{fig:concept} presents examples of 3D language understanding and novel-view generation concerning the same underlying scene properties.
In the top example, answering the question requires distinguishing the cabinet's color (\textbf{attributes}), which novel-view generation should also preserve when predicting the cabinet from a different viewpoint.
In the middle example, the question is about the stool's relative position (\textbf{spatial relationship}), while generation requires depicting this arrangement from the target viewpoint.
In the bottom example, identifying the chair as an armchair (\textbf{object category}) requires recognizing characteristic features that novel-view generation should preserve across viewpoints.
These examples suggest that learning to interpret scene properties through language can provide complementary supervision for novel-view generation.
We use this supervision during training to develop representations useful for generation, without requiring question answering when generating novel views.

%% file: sec/4_methods.tex
\input{figures/tex/arch}

\section{Methodology}
We introduce \textbf{\ours}, which strengthens semantic and spatial understanding for novel-view generation through supervision from 3D language understanding tasks.
Our model builds on Lance-Video-3B~\citep{fu2026lance}, a pretrained unified multimodal model with specialized generation and understanding transformers.
The generation transformer incorporates camera geometry into its visual inputs and enables bidirectional attention between clean source and noisy target VAE tokens (Sec.~\ref{cam_nvs}).
Representations from the understanding transformer are incorporated into novel-view generation through ViT conditioning and an attention mask that controls interactions among visual tokens (Sec.~\ref{und4gen}).
Then, we train the model with a flow-matching objective, supplemented by 3D language understanding objectives as auxiliary supervision (Sec.~\ref{train}).
We further analyze cross-view correspondence of VAE tokens and assess semantic and spatial knowledge of our model (Sec.~\ref{sec:analysis}).
The overall architecture is shown in Fig.~\ref{fig:Architecture}.

\input{sec/methodology/3_1_nvs}

\input{sec/methodology/3_3_enhancing}

\subsection{Training Objectives}
\label{train}
We train the model with the flow-matching objective for novel-view generation, supplemented by 3D question answering and visual grounding objectives as auxiliary supervision.
Training uses multimodal sequence packing, with attention blocked across independent samples.

The overall objective is
\begin{equation}
    \mathcal{L}_{\mathrm{total}}
    =
    \lambda_{\mathrm{gen}}\mathcal{L}_{\mathrm{gen}}
    +
    \lambda_{\mathrm{und}}\mathcal{L}_{\mathrm{und}},
    \qquad
    \mathcal{L}_{\mathrm{und}}
    =
    \mathcal{L}_{\mathrm{qa}}
    +
    \mathcal{L}_{\mathrm{grd}},
\end{equation}
where $\lambda_{\mathrm{gen}}$ and $\lambda_{\mathrm{und}}$ weight the generation and auxiliary understanding objectives.
The understanding objectives provide semantic and spatial supervision to the understanding transformer, whose representations are also used by the generation transformer through joint self-attention.

\input{sec/methodology/3_4_analysis}

%% file: figures/tex/arch.tex
\begin{figure*}[t]
    \centering
    \includegraphics[width=\linewidth]{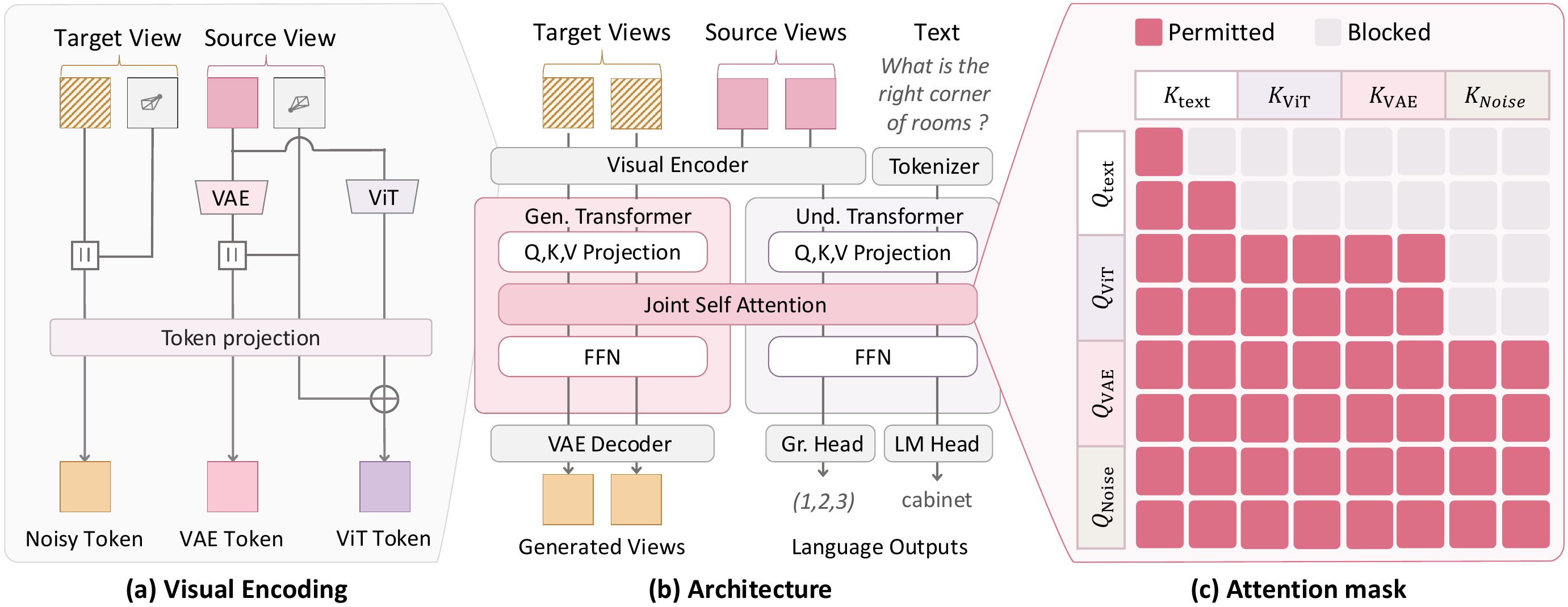}
    \vspace{-10pt}
    \caption{
        \textbf{Overall architecture of \ours.} 
        Given source images and camera parameters of the source and target views,
        (a) for novel-view generation, the images are first encoded into VAE and ViT features by their respective visual encoders, and camera geometry is incorporated into these representations before they are passed to the transformers.
        (b) The generation transformer processes the camera-aware VAE tokens, while the understanding transformer processes the camera-aware ViT tokens together with text tokens for 3D language understanding.
        (c) During novel-view generation, these modality-specific tokens interact through joint self-attention under a unified modality-aware attention mask.        
    }
    \label{fig:Architecture}
    \vspace{-10pt}
\end{figure*}

%% file: sec/methodology/3_1_nvs.tex
\subsection{Camera-Conditioned Novel-View Generation}
\label{cam_nvs}
Given $N$ source images and their corresponding camera parameters, the model must capture both scene appearance and geometric structure to generate $M$ target images at the specified target camera parameters.
The source images $\mathbf{I}_{\mathrm{src}}\in\mathbb{R}^{N\times H\times W\times 3}$ are encoded into VAE latents:
\begin{equation}
\mathbf{z}_{\mathrm{src}}
=
\mathcal{E}_{\mathrm{vae}}
\left(
\mathbf{I}_{\mathrm{src}}
\right)
\in
\mathbb{R}^{N\times h\times w\times C_z},
\qquad
h=\frac{H}{d_h},
\quad
w=\frac{W}{d_w},
\end{equation}
where $d_h$ and $d_w$ denote the spatial compression factors of the VAE and $C_z$ is the latent channel dimension.

We incorporate camera geometry into VAE features using Pl\"ucker rays.
Each ray is represented as $\mathbf{r}=[\mathbf{d};\mathbf{o}\times\mathbf{d}]\in\mathbb{R}^{6}$, where $\mathbf{o}$ and $\mathbf{d}$ denote the camera origin and ray direction.
We compute the rays for all pixels of the source images to form ray maps $\mathbf{R}_{\mathrm{src}}\in\mathbb{R}^{N\times H\times W\times 6}$ and rearrange them into the channel dimension to match the latent spatial resolution.
Since the VAE projector operates on channels while preserving the spatial grid, we extend its input channel dimension to accommodate the concatenated latent and ray features:
\begin{equation}
    \mathbf{R}_{\mathrm{src}}
    \in \mathbb{R}^{N\times H\times W\times 6}
    \;\longrightarrow\;
    \mathbf{z}_{\mathrm{src}}^{\mathrm{ray}}
    \in \mathbb{R}^{N\times h\times w\times C_r},
    \qquad
    \mathbf{e}^{\mathrm{vae}}
    =
    P_{\mathrm{vae}}
    \left(
    [\mathbf{z}_{\mathrm{src}};
    \mathbf{z}_{\mathrm{src}}^{\mathrm{ray}}]
    \right),
\end{equation}
where $C_r=6d_hd_w$ is the channel dimension of the rearranged ray representation.
We apply the same ray construction and projection to the target views, using their noisy VAE latents $\mathbf{z}_t\in\mathbb{R}^{M\times h\times w\times C_z}$ at timestep $t$:
\begin{equation}
    \mathbf{e}_t^{\mathrm{vae}}
    =
    P_{\mathrm{vae}}
    \left(
    [\mathbf{z}_t;
    \mathbf{z}_{\mathrm{tgt}}^{\mathrm{ray}}]
    \right),
\end{equation}
where $\mathbf{z}_{\mathrm{tgt}}^{\mathrm{ray}}\in\mathbb{R}^{M\times h\times w\times C_r}$ denotes the rearranged ray representations of the target views.

Whereas the pretrained model originally blocks attention from clean VAE tokens to noisy target tokens, we modify the attention mask to allow bidirectional attention between them, encouraging cross-view correspondence.

%% file: sec/methodology/3_3_enhancing.tex
\subsection{Enhancing Novel-view Generation through 3D Language Understanding}
\label{und4gen}
\input{tables/main_analysis}
We incorporate visual representations processed by the understanding transformer into novel-view generation, allowing generation to explicitly utilize semantic and spatial information.
Specifically, we extract ViT features from all source views and include them alongside the VAE tokens.
The understanding transformer processes these ViT tokens, and the generation transformer accesses their representations through joint self-attention.

Similar to the VAE representations, we incorporate camera information into the ViT features using Pl\"ucker ray embeddings.
We preserve the pretrained ViT projector's joint spatial merging and channel projection by adding ray embeddings to its output.
For each output visual token, we compute the ray at the center of its corresponding image region and map it into the language model's hidden space using a two-layer MLP:
\begin{equation}
    \tilde{\mathbf{e}}_{\mathrm{ray}}^{\mathrm{vit}}
    =
    \mathbf{e}^{\mathrm{vit}}
    +
    \mathrm{MLP}
    \left(
        \mathbf{r}^{\mathrm{vit}}
    \right),
\end{equation}
where $\mathbf{r}^{\mathrm{vit}}\in\mathbb{R}^{N_{\mathrm{vit}}\times 6}$ contains the rays corresponding to all source-view ViT tokens, and the MLP is applied independently to each ray.

We concatenate the camera-aware source VAE and ViT tokens, noisy target VAE tokens, and text tokens into an input sequence.
The two Transformers process their respective tokens through joint self-attention, with an attention mask controlling information flow among token groups.
Noisy target tokens can attend to both source VAE and ViT tokens, accessing source appearance and representations from the understanding transformer.
Source ViT tokens attend bidirectionally to clean VAE tokens but do not directly attend to noisy target tokens.
The complete attention pattern, including the bidirectional VAE attention described in Sec.~\ref{cam_nvs}, is illustrated in Fig.~\ref{fig:Architecture}(c).

The resulting target hidden states are projected to velocity predictions, and the model is trained with the flow-matching objective.
At inference, we initialize the target latents with Gaussian noise, integrate the predicted velocity field, and decode the resulting latents into target images using the VAE decoder.

%% file: tables/main_analysis.tex
\begin{table*}[t]

    \begin{minipage}[t]{\textwidth}
        \centering
        \includegraphics[width=\linewidth]{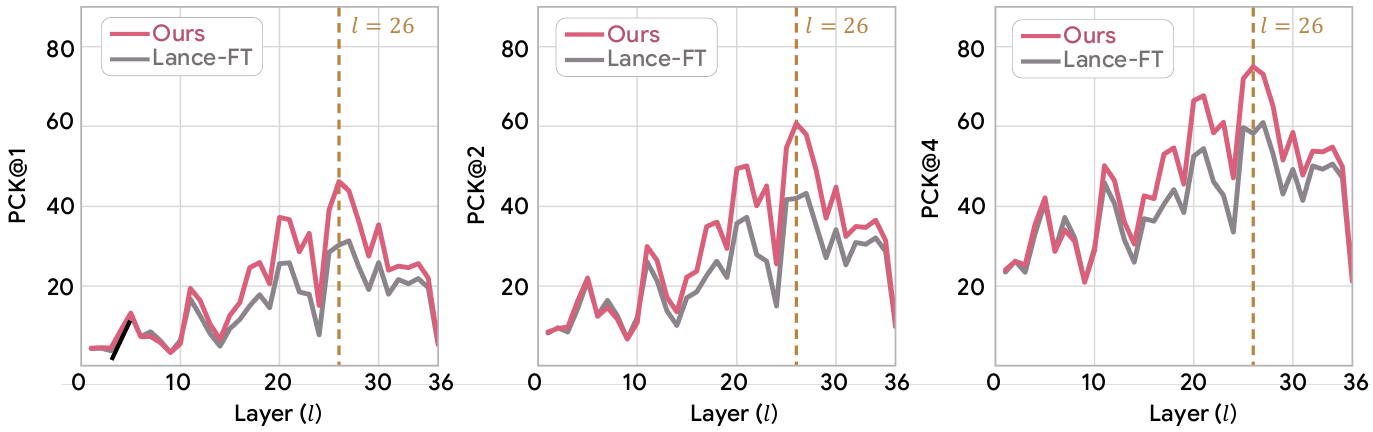}
        \vspace{-10pt}
        \captionof{figure}{
            \textbf{Layer-wise cross-view correspondence.} We report PCK@1, PCK@2, and PCK@4 across Transformer layers on ScanNet~\citep{dai2017scannet} for Lance-FT~\citep{fu2026lance} and \ours. Layer $l=26$ exhibits the strongest correspondence across all PCK thresholds.
        }
        \label{fig:corr_layer}
        \vspace{-10pt}
    \end{minipage}

\end{table*}

%% file: sec/methodology/3_4_analysis.tex
\section{Analyzing the Effects of Joint Training}
\label{sec:analysis}
In this section, we analyze how joint training with 3D language understanding benefits novel-view generation.
We compare \ours with two baselines initialized from the same pretrained Lance checkpoint~\citep{fu2026lance}.
Lance-FT is fine-tuned exclusively on novel-view generation, while Lance-FT + ViT uses the same ViT conditioning pathway and generation-side attention mask as Ask4NV, but is trained without 3D language understanding samples or objectives.
Both baselines are only trained on novel-view generation training samples.

\finding{3D language supervision improves cross-view correspondence beyond ViT conditioning.}{}
\input{tables/main_corr}
We examine whether adding 3D language understanding supervision improves cross-view correspondence in VAE representations, without explicitly supervising correspondence during training.
We evaluate correspondence in a two-view setting using clean VAE tokens from both views, focusing on corresponding tokens associated with the same 3D points across overlapping views.
Since correspondence quality varies across transformer layers, we first evaluate each layer independently.
Specifically, we extract the intermediate token feature after the attention computation and, for each object token in one view, identify its correspondence in the other view by selecting the token with the highest cosine similarity.
We evaluate the predicted correspondences using PCK@1, PCK@2, and PCK@4, where a prediction is considered correct if it lies within 1, 2, or 4 token-grid locations of the ground-truth correspondence.
As shown in Fig.~\ref{fig:corr_layer}, Layer $l=26$ achieves the strongest correspondence across the evaluated thresholds, and we therefore use its representations for the analysis.
Prior work such as VIRAL~\citep{yoon2025visual} highlights the importance of intermediate visual representations for fine-grained visual understanding.

Table~\ref{tab:corr} shows that \ours consistently outperforms both baselines across both datasets and all PCK thresholds.
Notably, although the additional understanding objectives supervise the understanding Transformer, correspondence improves in the generation Transformer's intermediate VAE representations without direct correspondence supervision.
This provides encouraging evidence that the benefits of 3D language understanding extend to spatial representations used for generation.

\finding{Joint training improves similarity in a 3D-aware feature space.}{}
\input{tables/main_preservation}
We analyze whether joint training improves the similarity between generated and ground-truth views in a 3D-aware representation space.
To this end, we use the SigLIP~\citep{zhai2023sigmoid} vision encoder fine-tuned for 3D language understanding within Video-3D-LLM~\citep{zheng2025video}.
We extract generated and ground-truth features at the same target viewpoint and compute the mean cosine similarity between spatially corresponding patch features.

Table~\ref{tab:vlm} shows that \ours achieves higher feature similarity than both generation-only baselines across all four datasets.
The improvement over Lance-FT + ViT indicates an additional benefit of joint training beyond ViT conditioning.
These results show that joint training produces generated views that are more closely aligned with the ground-truth views in the feature space of a 3D language understanding encoder.

%% file: tables/main_corr.tex
\begin{wraptable}{r}{0.5\textwidth}
    \vspace{-10pt}
    \centering

    \caption{
        \textbf{Cross-view Correspondence at Layer 26.}
        We report PCK@1, PCK@2, and PCK@4 on ScanNet and ScanNet++ for Lance-FT, Lance-FT + ViT and \ours.
    }
    \label{tab:corr}

    \vspace{-3pt}
    \small

    \resizebox{\linewidth}{!}{
        \begin{tabular}{c|l|ccc}
        \toprule
        Dataset
        & Method
        & PCK@1 & PCK@2 & PCK@4\\
        \midrule
        \multirow{3}{*}{ScanNet} & Lance-FT & 30.30 & 42.03 & 58.22\\
        & Lance-FT + ViT & 38.77 & 54.93 & 72.09\\
        & \cellcolor{red!10} \textbf{\ours (Ours)} & \cellcolor{red!10}\textbf{46.13} & \cellcolor{red!10} \textbf{60.73} &\cellcolor{red!10} \textbf{74.99}\\
        \midrule
        \multirow{3}{*}{ScanNet++} & Lance-FT & 33.34 & 44.31 & 60.28  \\
        & Lance-FT + ViT & 31.40 & 44.91 & 63.87\\
        &\cellcolor{red!10} \textbf{\ours (Ours)} & \cellcolor{red!10} \textbf{43.27} & \cellcolor{red!10} \textbf{55.25} & \cellcolor{red!10} \textbf{69.55} \\
        \bottomrule
        \end{tabular}
    }

    \vspace{-8pt}
\end{wraptable}

%% file: tables/main_preservation.tex
\begin{wraptable}{r}{0.5\textwidth}
    \vspace{-10pt}
    \centering

    \caption{
        \textbf{Patch-wise feature similarity between generated and ground-truth views.}
        We report feature similarity on DL3DV, ScanNet, RealEstate10K (Re10K), and ScanNet++.
    }
    \label{tab:vlm}

    \vspace{-3pt}
    \small

    \resizebox{\linewidth}{!}{
        \begin{tabular}{l|cccc}
        \toprule
        Method & DL3DV & ScanNet & Re10K & ScanNet++ \\
        
        \midrule
        Lance-FT & 0.512 & 0.498 & 0.580 & 0.514 \\
        Lance-FT + ViT & 0.546 & 0.555 & 0.636 & 0.538 \\
        \rowcolor{red!10} \textbf{\ours (Ours)} & \textbf{0.599} & \textbf{0.557} & \textbf{0.680} & \textbf{0.557} \\
        \bottomrule
        \end{tabular}
    }

    \vspace{-8pt}
\end{wraptable}

%% file: sec/5_experiments.tex
\section{Experiments}
\subsection{Experimental Setting}
\input{tables/main_quan_nvs}

\paragrapht{Implementation and training details.}
We build our model on Lance-Video-3B~\citep{fu2026lance}, a pretrained MoT-based unified multimodal model, and fine-tune it while freezing the ViT encoder and the VAE encoder and decoder.
Our 3D language understanding training data comprise ScanNet-based datasets, including SQA3D~\citep{ma2022sqa3d}, ScanQA~\citep{azuma2022scanqa}, ScanRefer~\citep{chen2020scanrefer}, and Multi3DRefer~\citep{zhang2023multi3drefer}.
We therefore use ScanNet~\citep{dai2017scannet} for novel-view generation to align the scene domains across the two tasks.
Given its limited number of scenes, we supplement the generation training data with DL3DV~\citep{ling2024dl3dv}.
We use multimodal sequence packing, with each pack containing three novel-view generation samples and one 3D language understanding sample.
This $3:1$ ratio approximates the pretraining task balance within our computational budget.
The understanding task uses $V=32$ uniformly sampled frames per scene.
For novel-view generation, we sample $8$ frames, using $2$--$4$ as source views and the remainder as target views.
Additional implementation details are provided in Appendix~\ref{sec:app_impl_detail}.

\input{figures/tex/qual_nvs}

\paragrapht{Evaluation protocol.}
To assess novel-view generation quality, we evaluate \ours on DL3DV~\citep{ling2024dl3dv} and ScanNet~\citep{dai2017scannet} as in-domain datasets, and RealEstate10K (Re10K)~\citep{zhou2018stereo} and ScanNet++~\citep{yeshwanth2023scannetpp} as out-of-domain datasets, relative to our training data.
In the main comparison, each model generates $M = 4$ target views from $N = 4$ source views.
We compare against diffusion-based novel view generation models, including CAT3D~\citep{gao2024cat3d}, Matrix3D~\citep{lu2025matrix3d}, NVComposer~\citep{li2025nvcomposer}, and CAMEO~\citep{kwon2025cameo}.
Omni-View~\citep{hu2026omni} is also included as a unified multimodal model baseline.
We report PSNR, SSIM, and LPIPS to measure 2D image fidelity, and camera estimation errors and MEt3R~\citep{asim2025met3r} to assess 3D geometric consistency.
Further details on evaluation are provided in Appendix~\ref{suppl:eval detail}.

\subsection{Main Results}
\finding{\ours improves novel-view image quality across multiple benchmarks.}{}
\paragrapht{Quantitative comparison of novel-view generation.}
As shown in Table~\ref{tab:3dnvs}, \ours achieves strong performance on 2D image quality metrics across multiple benchmarks, achieving the best PSNR and LPIPS on all four datasets while maintaining consistently strong SSIM across all benchmarks.
Our model also demonstrates consistently strong performance on 3D geometric metrics, achieving the best ATE across all datasets and strong results on $\mathrm{RPE}_r$, $\mathrm{RPE}_t$, and MEt3R.
Notably, the strong performance on the out-of-domain datasets, Re10K and ScanNet++, demonstrates the generalization ability of \ours.

\paragrapht{Qualitative comparison.}
Fig.~\ref{fig:nvs qual} provides a qualitative comparison of novel view generation between \ours and the baselines.
Given the source views, \ours generates novel views with sharper visual details while maintaining geometric consistency across diverse camera trajectories.
Additional qualitative results are provided in Appendix~\ref{supp: qual}.

\input{tables/main_quan_QA}
\subsection{Ablation and analysis}

\paragrapht{3D language understanding.}
We additionally evaluate 3D language understanding on SQA3D~\citep{ma2022sqa3d} and ScanQA~\citep{azuma2022scanqa} for question answering, and ScanRefer~\citep{chen2020scanrefer} and Multi3DRefer~\citep{zhang2023multi3drefer} for visual grounding.
Following prior work~\citep{zheng2025video, wang2025ross3d}, our model uses $V=32$ uniformly sampled frames per scene and is compared against 2D LMMs and 3D LLMs.
Tab.~\ref{tab:3dqa} presents the quantitative results.
These results show that \ours acquires 3D language understanding capabilities alongside its primary novel-view generation objective.

\input{tables/main_source_num}
\paragrapht{Novel-view generation on different number of source views.}
Tab.~\ref{tab:view} presents novel-view generation results on DL3DV with $N=2$ and $N=3$ source views, predicting $M=6$ and $M=5$ target views, respectively.
\ours consistently outperforms Lance-FT, showing that the benefit of joint training persists under fewer source views.

\input{tables/main_attn}
\paragrapht{Ablation on bidirectional VAE attention.}
As shown in Tab.~\ref{tab:att}, bidirectional attention between clean source and noisy target VAE tokens improves performance on DL3DV, demonstrating the benefit of direct source-target interaction during generation.

%% file: tables/main_quan_nvs.tex
\begin{table*}[t]
    \caption{
    \textbf{Evaluation of novel view generation.} We evaluate on DL3DV and ScanNet (in-domain for \ours) and ScanNet++ and Re10K (out-of-domain for \ours).
    We report metrics for both 2D image fidelity and 3D geometric consistency.
    $\dagger$ denotes the reproduced CAT3D model.
    \textbf{Bold} and \underline{underline} indicate the best and second-best, respectively.
    }
    \label{tab:3dnvs}
    \centering\small
    \resizebox{\linewidth}{!}{
    \begin{tabular}{c|l|ccc|cccc}
    \toprule
    \multirow{2}{*}{Dataset} & \multirow{2}{*}{Method} & \multicolumn{3}{c|}{\textbf{2D Metrics}} & \multicolumn{4}{c}{\textbf{3D Metrics}} \\
    & & PSNR $\uparrow$ & SSIM $\uparrow$ & LPIPS $\downarrow$ & ATE $\downarrow$ & RPE$\mathrm{_r}$ $\downarrow$ & RPE$\mathrm{_t}$ $\downarrow$ & MEt3R $\downarrow$ \\
    \midrule
    \multirow{7}{*}{DL3DV} & CAT3D$^{\dagger}$~\citep{gao2024cat3d} & 12.662 & 0.287 & 0.489 & 0.777 & 15.59 & 1.674 & 0.446 \\
    & Matrix3D~\citep{lu2025matrix3d} & \underline{13.997} & \underline{0.344} & \underline{0.412} &
    0.351 & 4.80 & 0.691 & \underline{0.358}	\\
    & NVComposer~\citep{li2025nvcomposer} & 11.532 & 0.255 & 0.560 & 0.995 & 32.34 & 1.748 & 0.511\\
    & CAMEO~\citep{kwon2025cameo} & 12.991 & 0.312 & 0.463 & 0.628 & 13.16 & 1.174 & 0.388\\
    & Omni-View~\citep{hu2026omni} & 13.009 & 0.221 & 0.558 & 1.226 & 27.86 & 2.514 & 0.511	\\
    & Lance-FT~\citep{fu2026lance}& 13.291 & 0.297 & 0.432 & \underline{0.185} & \underline{3.80} & \underline{0.354} & 0.361 \\
    & \cellcolor{red!10} \textbf{\ours \ (Ours)} & \cellcolor{red!10} \textbf{14.939} &  \cellcolor{red!10} \textbf{0.355} & \cellcolor{red!10} \textbf{0.313} & \cellcolor{red!10} \textbf{0.158} & \cellcolor{red!10} \textbf{3.11} & \cellcolor{red!10} \textbf{0.298} & \cellcolor{red!10} \textbf{0.349}\\
    \midrule    
    \multirow{7}{*}{ScanNet} & CAT3D$^{\dagger}$~\citep{gao2024cat3d} & 12.007 & 0.438 & 0.627 &0.366 & 33.12& 0.719 & 0.559\\
    & Matrix3D~\citep{lu2025matrix3d} & \underline{14.298} & \underline{0.539} & \underline{0.498} & \underline{0.212} & 20.37 & 0.431 & 0.417\\
    & NVComposer~\citep{li2025nvcomposer} & 10.186 & 0.368 & 0.709 & 0.443 & 64.69 & 0.787 & 0.720\\
    & CAMEO~\citep{kwon2025cameo} & 12.417 & 0.485 & 0.605 & 0.343 & 32.86 & 0.734 & 0.492\\
    & Omni-View~\citep{hu2026omni} & 12.380 & 0.429 & 0.664 & 0.469 & 41.61 & 0.958 & 0.653\\
    & Lance-FT~\citep{fu2026lance} & 14.292 & 0.530 & 0.518 & 0.233 & \underline{15.45} & \underline{0.351} & \textbf{0.398}\\
    & \cellcolor{red!10} \textbf{\ours \ (Ours)} & \cellcolor{red!10} \textbf{15.747} & \cellcolor{red!10} \textbf{0.564} & \cellcolor{red!10} \textbf{0.434} & \cellcolor{red!10} \textbf{0.191 }& \cellcolor{red!10} \textbf{13.85} & \cellcolor{red!10} \textbf{0.340} & \cellcolor{red!10} \underline{0.413}\\
    \midrule
    \multirow{7}{*}{Re10K} & CAT3D$^{\dagger}$~\citep{gao2024cat3d} & 16.261 & 0.555 & 0.330 & 0.168 & 3.25 & 0.317 & 0.345\\
    & Matrix3D~\citep{lu2025matrix3d} & 16.374 & \underline{0.566} & 0.319 & 0.154& 1.96 & 0.298 & \textbf{0.303}\\
    & NVComposer~\citep{li2025nvcomposer} & 12.716 & 0.439 & 0.551 & 0.507 & 21.40 & 0.819 &0.416\\
    & CAMEO~\citep{kwon2025cameo} & \underline{16.664} & \textbf{0.577} & \underline{0.301} & 0.121 & 3.07 & 0.223 & \underline{0.312}\\
    & Omni-View~\citep{hu2026omni} & 13.408 & 0.371 & 0.566 & 0.871 & 18.82 & 1.657 & 0.536\\
    & Lance-FT~\citep{fu2026lance} & 14.548 & 0.494 & 0.432 & \underline{0.116} & \underline{1.91} & \underline{0.221} & 0.328\\
    & \cellcolor{red!10}\textbf{\ours \ (Ours)} & \cellcolor{red!10} \textbf{16.738} & \cellcolor{red!10} 0.562 & \cellcolor{red!10} \textbf{0.290} & \cellcolor{red!10} \textbf{0.106} & \cellcolor{red!10}\textbf{1.68} & \cellcolor{red!10}\textbf{0.202} & \cellcolor{red!10}0.319\\
    \midrule
    \multirow{7}{*}{ScanNet++} & CAT3D$^{\dagger}$~\citep{gao2024cat3d} & 11.940 & 0.475 & 0.627 & 0.480 & 54.42 & 1.138 & 0.572\\
    & Matrix3D~\citep{lu2025matrix3d} & \underline{14.667} & \textbf{0.587} & \underline{0.453} & \textbf{0.287} & \textbf{14.19} & 0.661 & \textbf{0.375}\\
    & NVComposer~\citep{li2025nvcomposer} & 10.725 & 0.446 & 0.653 & 0.604 & 76.51 & 1.210 & 0.626\\
    & CAMEO~\citep{kwon2025cameo} & 12.544 & 0.532 & 0.590 & 0.511 & 48.86 & 1.046 & 0.433\\
    & Omni-View~\citep{hu2026omni} & 12.536 & 0.457 & 0.654 & 0.641 & 65.43 & 1.375 & 0.667\\
    & Lance-FT~\citep{fu2026lance} & 14.468 & 0.571 & 0.483 &  \underline{0.301} & 19.58 & \textbf{0.612} & 0.417\\
    & \cellcolor{red!10}\textbf{\ours \ (Ours)} & \cellcolor{red!10}\textbf{15.242} & \cellcolor{red!10} \underline{0.583} & \cellcolor{red!10} \textbf{0.419} & \cellcolor{red!10} \textbf{0.287} & \cellcolor{red!10} \underline{16.54} & \cellcolor{red!10} \underline{0.649} & \cellcolor{red!10} \underline{0.398}\\
    \bottomrule
    \end{tabular}
    }
    \vspace{-5pt}

\end{table*}

%% file: figures/tex/qual_nvs.tex
\begin{figure*}[t]
    \centering
    \includegraphics[width=\linewidth]{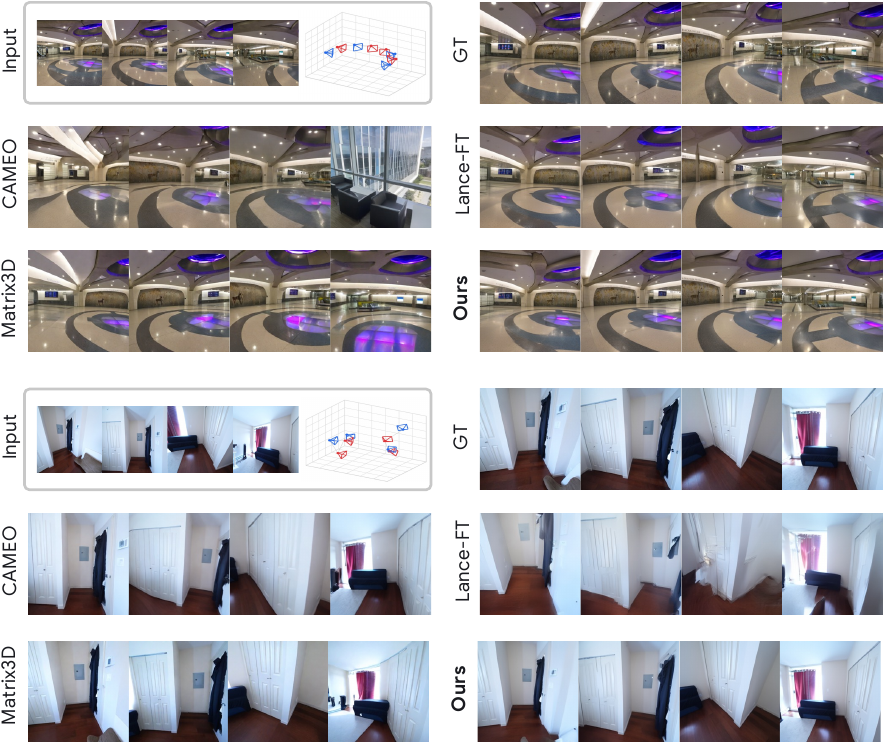}
    \vspace{-10pt}
    \caption{
        \textbf{Qualitative results of novel view generation} of \ours and baselines on in-domain DL3DV~\citep{ling2024dl3dv} and out-of-domain ScanNet++~\citep{yeshwanth2023scannetpp}. \textcolor{blue}{Blue} and \textcolor{red}{red} cameras indicate source and target viewpoints, respectively.
    }
    \label{fig:nvs qual}
    \vspace{-10pt}
\end{figure*}

%% file: tables/main_quan_QA.tex
\begin{table*}[t]
    \caption{
    \textbf{Evaluation of 3D language understanding.} We provide quantitative comparison of \ours and baselines on 3D question answering (SQA3D~\citep{ma2022sqa3d} and ScanQA~\citep{azuma2022scanqa}) and 3D visual grounding (ScanRefer~\citep{chen2020scanrefer} and Multi3DRefer~\citep{zhang2023multi3drefer}). Results are reported from prior works, and ``-" indicates an unavailable result.
    C, B-4, M, and R denote CIDEr, BLEU-4, METEOR, and ROUGE, respectively.
    }
    \centering\small
    \resizebox{\textwidth}{!}{
    \begin{tabular}{lcccccccccccc}
    \toprule
    \multirow{2}{*}{Method} & \multirow{2}{*}{Size} & \multicolumn{2}{c}{SQA3D$_{\text{test}}$}& \multicolumn{5}{c}{ScanQA$_{\text{val}}$} & \multicolumn{2}{c}{ScanRefer$_{\text{val}}$} & \multicolumn{2}{c}{Multi3DRefer$_{\text{val}}$} \\
    \cmidrule(lr){3-4}
    \cmidrule(lr){5-9}
    \cmidrule(lr){10-11}
    \cmidrule(lr){12-13}
    & & EM & EM-R & C & B-4 & M & R & EM & Acc@0.25 & Acc@0.5 & F1@0.25 & F1@0.5\\
    \midrule\midrule
    \multicolumn{13}{l}{\textcolor{gray}{\textit{2D LMMs}}} \\
    InternVL2-8B~\citep{chen2024internvl} & 8B & 33.0 & 45.3 & 62.5 & 3.3 & 14.5 & 34.3 & -- & -- & -- & -- & -- \\
    Qwen2.5-VL-3B~\citep{Qwen2.5-VL} & 3B & 46.3 & 49.0 & 67.7 & 9.7 & 12.9 & 35.6 & 23.9 & -- & -- & -- & -- \\
    LLaVA-Video-7B~\citep{zhang2024llava} & 7B & 48.5 & -- & 88.7 & 3.1 & 17.7 & 44.6 & -- & -- & -- & -- & -- \\
    Lance-3B~\citep{fu2026lance} & 3B & 23.4 & 37.3 & 6.0 & 0.3 & 11.4 & 9.9 & 0.3 & -- & -- & -- & -- \\
    \midrule
    \multicolumn{13}{l}{\textcolor{gray}{\textit{3D LMMs}}} \\
    LLaVA-3D~\citep{zhu2024llava} & 7B & 55.6 & 57.6 & 91.7 & 14.5 & \underline{20.7} & \underline{50.1} & 27.0 & 54.1 & 42.4 & -- & -- \\
    Video-3D-LLM~\citep{zheng2025video} & 7B & 58.6 & -- & 102.1 & \underline{16.4} & 20.0 & 49.3 & 30.1 & \underline{58.1} & \underline{51.7} & \underline{58.0} & \underline{52.7} \\
    3DRS~\citep{huang20263drs} & 7B & \underline{60.6} & -- & \underline{104.8} & -- & -- & -- & \underline{30.3} & -- & -- & -- & --\\
    Ross3D~\citep{wang2025ross3d} & 7B & \textbf{63.0} & \textbf{65.7} & \textbf{107.0} & \textbf{17.9} & \textbf{20.9} & \textbf{50.7} & \textbf{30.8} & \textbf{61.1} & \textbf{54.4} & \textbf{59.6} & \textbf{54.3} \\
    Spatial-MLLM~\citep{wu2026spatial} & 3B & 55.9 & \underline{58.7} & 91.8 & 14.8 & 18.4 & 45.0 & -- & -- & -- & -- & -- \\
    Omni-View~\citep{hu2026omni} & 7B & 59.2 & 61.9 & 103.0 & 16.2 & 20.1 & 49.0 & 29.5 & 50.8 & 45.0 & -- & -- \\
    \rowcolor{red!10} \textbf{\ours \ (Ours)} & 3B  & 54.9 & 57.7 & 87.7 & 13.6 & 17.4 & 43.7 & 25.6 & 52.9 & 47.3 & 52.6 & 48.3 \\
    \bottomrule
    \end{tabular}
    }
    \vspace{-5pt}

    \label{tab:3dqa}
\end{table*}

%% file: tables/main_source_num.tex
\begin{wraptable}[8]{r}{0.5\textwidth}
    \vspace{-15pt}
    \centering

    \captionof{table}{
    \textbf{Ablation on different numbers of source views.}
    }
    \label{tab:view}
    \centering\small
    \resizebox{\linewidth}{!}{
        \begin{tabular}{c|l|ccc}
        \toprule
        Source & \multirow{2}{*}{Method} & \multirow{2}{*}{PSNR $\uparrow$} & \multirow{2}{*}{SSIM $\uparrow$} & \multirow{2}{*}{LPIPS $\downarrow$}\\
        Views & & & &\\
        \midrule
        \multirow{2}{*}{$N=2$} & Lance-FT & 13.001 & 0.294 & 0.459 \\
        &  \cellcolor{red!10}\textbf{\ours (Ours)} & \cellcolor{red!10} \textbf{13.900} & \cellcolor{red!10} \textbf{0.329} & \cellcolor{red!10} \textbf{0.388} \\
        \midrule
        \multirow{2}{*}{$N=3$} & Lance-FT & 13.292 & 0.298 & 0.430 \\
        & \cellcolor{red!10} \textbf{\ours (Ours)} & \cellcolor{red!10} \textbf{14.559} & \cellcolor{red!10} \textbf{0.344} &\cellcolor{red!10} \textbf{0.355} \\
        \bottomrule
    \end{tabular}
    }

    \vspace{-8pt}
\end{wraptable}

%% file: tables/main_attn.tex
\begin{wraptable}[7]{r}{0.5\textwidth}
    \vspace{-10pt}
    \centering

    \captionof{table}{\textbf{Ablation on bidirectional VAE attention.}}
    \label{tab:att}
    \resizebox{\linewidth}{!}{
        \begin{tabular}{l|ccc}
        \toprule
        Method & PSNR $\uparrow$ & SSIM $\uparrow$ & LPIPS $\downarrow$ \\
        \midrule
        Unidirectional & 13.653  & 0.314 & 0.424 \\
        \rowcolor{red!10} \textbf{Bidirectional (Ours)} & \textbf{14.939}  & \textbf{0.355} & \textbf{0.313} \\
        \bottomrule
        \end{tabular}
        }

    \vspace{-8pt}
\end{wraptable}

%% file: sec/6_conclusions.tex
\section{Conclusions}
In this work, we investigate how joint training with 3D language understanding can enhance novel-view generation.
We introduce \ours, which adapts a pretrained MoT-based architecture through ViT conditioning and a tailored attention mask.
Experiments demonstrate improvements in image fidelity and geometric consistency across multiple benchmarks, while maintaining competitive 3D language understanding performance.
Our analyses show improved cross-view correspondence in VAE representations and stronger alignment with ground-truth views in a 3D-aware feature space.
Ablation studies further validate the benefits of joint training and our architectural choices.
These findings highlight the potential of 3D language understanding as complementary supervision for generative scene modeling.

\subsection*{AI Use Statement}
We used generative AI tools to assist with language editing to improve readability and clarity.
We did not use generative AI tools for generating synthetic datasets, developing theoretical models, formulating or proving mathematical claims, proposing hypothesis, designing experiments, or interpreting results.
The authors reviewed and revised the AI-assisted material and take responsibility for the final content of this work, including its text, claims, and artifacts produced with AI assistance.

\subsection*{Ethics Statement}
This work investigates novel-view generation and 3D language understanding using existing publicly available datasets.
We do not identify additional ethical concerns specific to the proposed methodology beyond those generally associated with visual generation, including the potential misuse of synthetic images.

\subsection*{Reproducibility Statement}
We provide the model formulation and training objectives in the main paper, with additional details to support reproduction of our experiments in the supplementary material.
Sec.~\ref{sec:app_impl_detail} describes the training configuration, architectural components, and attention masks.
Sec.~\ref{supp:3d_und} details the implementation of 3D language understanding, including positional embeddings and the visual grounding objective.
Evaluation metrics and baseline settings are provided in Sec.~\ref{suppl:eval detail}.

%% file: sec/X_appendix.tex
\clearpage
\appendix
\etocdepthtag.toc{appendix} 
\section*{\Large Appendix}

\makeatletter
\newcommand{\appendix@noopaddcontentsline}[3]{}
\let\appendix@origsection\section
\renewcommand{\section}{\@ifstar\appendix@origsection\appendix@section}
\newcommand{\appendix@section}[1]{%
  \let\appendix@savedaddcontentsline\addcontentsline
  \let\addcontentsline\appendix@noopaddcontentsline
  \appendix@origsection{#1}%
  \let\addcontentsline\appendix@savedaddcontentsline
  \phantomsection
  \addcontentsline{toc}{section}{\protect\numberline{\thesection}#1}%
}
\let\appendix@origsubsection\subsection
\renewcommand{\subsection}{\@ifstar\appendix@origsubsection\appendix@subsection}
\newcommand{\appendix@subsection}[1]{%
  \let\appendix@savedaddcontentsline\addcontentsline
  \let\addcontentsline\appendix@noopaddcontentsline
  \appendix@origsubsection{#1}%
  \let\addcontentsline\appendix@savedaddcontentsline
  \phantomsection
  \addcontentsline{toc}{subsection}{\protect\numberline{\thesubsection}#1}%
}
\makeatother

\etocsetlevel{author}{6}
\etocsetlevel{title}{6}
\etocsettagdepth{main}{none}
\etocsettagdepth{appendix}{subsection}
\makeatletter
{%
\let\section\appendix@origsection
\let\subsection\appendix@origsubsection
\tableofcontents
}
\makeatother

\etocdepthtag.toc{appendix} 

\clearpage
\input{sec/appendix/X_A_impl_detail}

\newpage
\input{sec/2_related_works}

\input{sec/appendix/X_C}

\section{Limitations and Future work}
Our experiments demonstrate improvements in novel-view generation under the evaluated multi-view settings.
Extending these benefits to longer camera trajectories and more diverse in-the-wild scenes remains an important direction, requiring consistent preservation of scene appearance and geometry across broader viewing conditions.
Our current training and evaluation also use a fixed image resolution.
Supporting multiple resolutions and higher-resolution outputs would broaden the method's applicability while introducing additional challenges in preserving fine visual details across views.
Larger models and more diverse training data would enable further investigation of how the benefits of 3D language supervision scale.
Although a performance gap remains relative to leading 3D language understanding models, future unified backbones with stronger understanding capabilities may help narrow this gap and provide richer supervision for novel-view generation.

\newpage
\input{figures/tex/supp_qual_dl3dv_s4}

\input{figures/tex/supp_qual_scannet_s4}

\input{figures/tex/supp_qual_re10k_s4}

\input{figures/tex/supp_qual_pp_4}

\input{figures/tex/supp_source_3}

\input{figures/tex/supp_source_2}

%% file: sec/appendix/X_A_impl_detail.tex
\section{Additional Implementation Details}
\subsection{Dataset Details}

\paragrapht{Novel view generation.}
We use ScanNet~\citep{dai2017scannet} for both training and evaluation, as our 3D language understanding datasets are also based on ScanNet.
Given the limited number of scenes in ScanNet, we supplement the training data with DL3DV~\citep{ling2024dl3dv} to increase scene diversity and reduce the risk of overfitting.
We additionally evaluate on RealEstate10K~\citep{zhou2018stereo} and ScanNet++~\citep{yeshwanth2023scannetpp}, neither of which is used for training, to assess cross-dataset generalization.

\paragrapht{3D language understanding.}
We use the preprocessed ScanNet-based datasets provided by Video-3D-LLM~\citep{zheng2025video}, including SQA3D~\citep{ma2022sqa3d}, ScanQA~\citep{azuma2022scanqa}, ScanRefer~\citep{chen2020scanrefer}, and Multi3DRefer~\citep{zhang2023multi3drefer}.

\paragrapht{Camera normalization.}
For both tasks, we express camera poses relative to the first input camera and normalize camera centers by their mean pairwise distance. For novel-view generation, this distance is computed over both source and target camera centers.
We apply the same coordinate transformation and scale normalization to the associated 3D points and object boxes.

\subsection{3D Language Understanding Details}
\label{supp:3d_und}

\paragrapht{3D coordinate-aware visual representations.}
Given $V$ input images and their associated pixel-wise 3D coordinates,
we encode the ViT representations with explicit 3D positional embeddings, as in prior work~\citep{zheng2025video, wang2025ross3d}.
For each visual token $k$, we average the 3D coordinates over its corresponding pixel region $\mathcal{P}_k$ and add the sinusoidal encoding of the resulting coordinate to its visual feature:
\begin{equation}
    \mathbf{p}_k
    =
    \frac{1}{|\mathcal{P}_k|}
    \sum_{\mathbf{u}\in\mathcal{P}_k}
    \mathbf{p}(\mathbf{u}),
    \qquad
    \tilde{\mathbf{e}}_k^{\mathrm{vit}}
    =
    \mathbf{e}_k^{\mathrm{vit}}
    +
    \phi_{\mathrm{3D}}(\mathbf{p}_k),
\end{equation}
\label{eq:3d}
where $\mathbf{p}(\mathbf{u})$ is the 3D point obtained by back-projecting pixel $\mathbf{u}$ using the dataset-provided depth and camera intrinsics and transforming it into a common coordinate frame using the camera pose.
The function $\phi_{\mathrm{3D}}$ denotes sinusoidal positional encoding~\citep{vaswani2017attention}.
Here, $k$ indexes visual tokens across all input views, and each
$\mathcal{P}_k$ accounts for the spatial merging performed by the
pretrained ViT projector.
The resulting sequence $\tilde{\mathbf{e}}_{\mathrm{3D}}^{\mathrm{vit}} = (\tilde{\mathbf{e}}_k^{\mathrm{vit}})_k$ is combined with the system prompt and task instruction embeddings and processed by the understanding transformer.

These depth-derived 3D positional embeddings are used only for 3D language understanding samples and are not used for novel-view generation.

\paragrapht{Question answering.}
The understanding transformer autoregressively generates an answer from
the spatially enriched visual tokens and the input question.
We apply the cross-entropy objective $\mathcal{L}_{\mathrm{qa}}$ to the ground-truth answer tokens.

\paragrapht{Visual grounding.}
We adopt the proposal-based grounding formulation used in prior
3D LLMs~\citep{zheng2025video, wang2025ross3d}.
A dedicated grounding token is placed at the response position,
and its final-layer hidden state $\mathbf{h}_{\mathrm{grd}}$
serves as the query representation for matching object proposals.

Each object representation $\mathbf{e}_l^{\mathrm{obj}}$ is constructed
by mean-pooling the associated ViT features before token-level
3D positional encoding is added, and then adding the 3D positional
encoding of the object center.
We augment the candidate set with a learned no-object embedding
to handle descriptions with no matching objects.

The matching score for candidate $l$ and the grounding loss are
defined as
\begin{equation}
    s_l
    =
    \frac{
        f(\mathbf{e}_l^{\mathrm{obj}})^{\top}
        g(\mathbf{h}_{\mathrm{grd}})
    }{\tau_{\mathrm{grd}}},
    \qquad
    \mathcal{L}_{\mathrm{grd}}
    =
    -\log
    \frac{
        \sum_{l\in\mathcal{O}^{+}}\exp(s_l)
    }{
        \sum_{l\in\tilde{\mathcal{O}}}\exp(s_l)
    }.
\end{equation}
Here, $\tilde{\mathcal{O}}$ denotes the index set of all candidates,
including the no-object candidate, and $\mathcal{O}^{+}$ contains
the indices of the ground-truth positive candidates.
The functions $f$ and $g$ are two-layer MLPs followed by
$\ell_2$ normalization, and $\tau_{\mathrm{grd}}$ is the
temperature hyperparameter.

During training, we construct the object candidate set from ground-truth object boxes and use the annotated referents as positive candidates.
For descriptions with no matching objects, the no-object candidate is the sole positive.
At evaluation, we replace ground-truth boxes with Mask3D object proposals and score the candidates using the same grounding head.
Candidates are ranked in descending order of their softmax probabilities, and we select the smallest set whose cumulative probability mass reaches at least $0.4$.
For Multi3DRefer, predicted and ground-truth boxes are optimally matched within each query based on 3D IoU.
The resulting matches are used to compute precision, recall, and F1 at IoU thresholds of $0.25$ and $0.5$.

\subsection{Training and Architecture Details.}
\label{sec:app_impl_detail}

\paragrapht{Training configuration.}
We train our model using the Adam8bit optimizer~\citep{dettmers20218} on $4$ A100 GPUs, and accumulate gradients over $64$ micro-steps per optimizer update.
We train for 728 optimizer steps, corresponding to one epoch over the 3D language understanding data, with a linear warmup over 40 steps to $5\times10^{-5}$, followed by a constant learning rate (held at $5\times10^{-5}$).
For novel-view generation, we use a timestep shift of $4.0$ and apply conditioning dropout with a probability of $0.1$ to each of the text and camera conditions.
For 3D visual grounding, we set the temperature hyperparameter $\tau_{\mathrm{grd}} = 0.07$.
Following our backbone, we set the loss weights $\lambda_{\mathrm{gen}}$ and $\lambda_{\mathrm{und}}$ to $1.0$ and $0.25$, respectively.
For a fair comparison, Lance-FT, Lance-FT + ViT, and Ask4NV are trained for the same number of optimizer steps and observe the same number of novel-view generation samples.

\paragrapht{Architectural details.}
We expand the input dimension of the pretrained projector $P_{\mathrm{vae}}$ from $C_z$ to $\tilde{C}_z=C_z+C_r$ to accommodate the concatenated ray representations.
We retain the pretrained weights for the latent channels and zero-initialize the weights for the additional ray channels.
We resize input images to $480\times480$ for the VAE encoder and $476\times476$ for the ViT encoder, ensuring compatibility with the VAE's spatial compression factor of $16$ and the ViT's patch size of $14\times14$.
We apply temporal padding for VAE encoding to eliminate the effect of its temporal compression while maintaining a correspondence between input views and latent representations.

\subsection{Evaluation details.}
\label{suppl:eval detail}

\paragrapht{Camera-pose accuracy.}
In addition to PSNR, SSIM, and LPIPS, we evaluate camera-pose accuracy and multi-view consistency.
For camera-pose evaluation, we jointly estimate poses from the generated target views using a fixed VGGT-$\Omega$-1B-512 checkpoint~\citep{wang2026vggtomega}, without including source images or ground-truth target images.
We align the estimated camera centers to the corresponding ground-truth centers using a least-squares Sim(3) transformation computed with the Umeyama algorithm.

We report Absolute Trajectory Error (ATE) as the root mean squared error (RMSE) of the aligned target camera centers.
We also compute Relative Pose Errors for rotation ($\mathrm{RPE}_{r}$) and translation ($\mathrm{RPE}_{t}$) between consecutive target views, reusing the same Sim(3) alignment.
Rotation error is measured by the geodesic angle in degrees, and translation error by the Euclidean norm of the translation component of the relative-pose error.
We report RMSE across the evaluated pairs, with ATE and $\mathrm{RPE}_{t}$ expressed in ground-truth coordinate units.

\paragrapht{Multi-view consistency.}
We evaluate multi-view consistency using a variant of MEt3R~\citep{asim2025met3r} that replaces its original point-map reconstruction backbone with VGGT-$\Omega$, using the same configuration for all evaluated methods.
For each pair of generated views $I_a$ and $I_b$, we extract dense DINO features~\citep{caron2021emerging} and upsample them with FeatUp~\citep{fu2024featup}.
Using the reconstructed point maps, we reproject and render both feature maps into a common viewpoint at $256\times256$ resolution.
We compute their mean pixel-wise cosine similarity over the valid overlapping region in both projection directions:
\begin{equation}
    \mathrm{MEt3R}(I_a,I_b)
    =
    1-
    \frac{
        S_{a\leftarrow b}
        +
        S_{b\leftarrow a}
    }{2},
\end{equation}
where $S_{a\leftarrow b}$ denotes the similarity between the two feature maps rendered in the viewpoint of $I_a$.
A lower score indicates stronger multi-view consistency.

We evaluate all unordered pairs of generated target views and average their scores, retaining only pairs with valid scores in both directions. For all geometry metrics, we first average valid tuple-level scores within each scene and then average across scenes, giving equal weight to each scene.

\paragrapht{3D language understanding metrics.}
We report EM and EM-R~\citep{huang2023embodied, ma20263d, zhu2024llava} on SQA3D, and EM, CIDEr~\citep{vedantam2015cider}, BLEU-4~\citep{papineni2002bleu}, METEOR~\citep{banerjee2005meteor}, and ROUGE~\citep{lin2004rouge} on ScanQA.
For visual grounding, we report Acc@0.25 and Acc@0.5 on ScanRefer, and F1@0.25 and F1@0.5 on Multi3DRefer, where the thresholds specify the 3D bounding-box IoU.

\paragrapht{Setting of baselines.}
The evaluated baselines differ in their input requirements, including their reliance on external modules for geometric estimation or number of source or target views.
For a fair comparison on our benchmarks, we follow a common evaluation protocol while preserving the conditioning requirements of each method.
Below, we describe the method-specific preprocessing and inference settings used for evaluation.

We focus our comparison on methods that generate novel views without explicit depth-based warping or point-cloud rendering as inference-time conditioning.
Accordingly, we exclude MVGenMaster~\citep{cao2025mvgenmaster}, GEN3C~\citep{ren2025gen3c}, and ViewCrafter~\citep{yu2024viewcrafter}, which use such geometric guidance during generation.

\textit{CAT3D}~\citep{gao2024cat3d}: Since the official implementation is not publicly available, we use the reproduction provided by CAMEO~\citep{kwon2025cameo}.
This reproduction is trained with camera parameters at the scale of VGGT~\citep{wang2025vggt}, so we use VGGT-estimated camera parameters during evaluation as well.
As the model is trained exclusively on Re10K, we treat DL3DV, ScanNet, and ScanNet++ as out-of-domain evaluation datasets.

\textit{NVComposer}~\citep{li2025nvcomposer}: We use source images at $512\times512$ resolution and ground-truth camera parameters.
Since the model processes a fixed set of 16 views, we fill the remaining slots with cameras interpolated between the first source view and the target views, and evaluate only the target views. 
As ScanNet and ScanNet++ are unseen during NVComposer's training, we treat them as out-of-domain evaluation datasets.

\textit{CAMEO}~\citep{kwon2025cameo}: As CAMEO shares the architecture and inference pipeline of the CAT3D reproduction, we likewise use camera parameters estimated by VGGT~\citep{wang2025vggt} during evaluation. 
It is also trained exclusively on Re10K, so we treat DL3DV, ScanNet, and ScanNet++ as out-of-domain evaluation datasets.

\textit{Matrix3D}~\citep{lu2025matrix3d}: The model generates the target images in a single pass, conditioned on the source images and the ground-truth camera parameters of all views. It is trained on Objaverse~\citep{deitke2023objaverse}, MVImgNet~\citep{yu2023mvimgnet}, CO3D-v2~\citep{reizenstein21co3d}, Re10K~\citep{zhou2018stereo}, Hypersim~\citep{roberts2021hypersim}, and ARKitScenes~\citep{baruch2021arkitscenes}. We therefore treat DL3DV, ScanNet, and ScanNet++ as out-of-domain evaluation datasets.

\textit{Omni-View}~\citep{hu2026omni}: Omni-View is a unified multiview 3D model that couples understanding, texture (appearance), and geometry modules. 
Novel views are synthesized by its texture module. 
Its novel-view generation training uses only Re10K.
Thus, DL3DV, ScanNet, and ScanNet++ are outside its generation training distribution, although ScanNet-based datasets are used for understanding training.
We evaluate its texture module using four source views at $512\times512$ resolution and the ground-truth source and target camera parameters.

%% file: sec/2_related_works.tex
\section{Related Work}
\subsection{Novel View Generation}
Novel-view synthesis (NVS) aims to predict the appearance of a scene from unseen viewpoints given observed views.
Reconstruction-based approaches, such as NeRF~\citep{mildenhall2021nerf} and 3D Gaussian Splatting~\citep{kerbl20233d}, model scenes using 3D representations from which novel views are rendered.
Recent advances in generative modeling~\citep{ho2020denoising, rombach2022high, peebles2023scalable} have enabled novel-view generation from sparse observations by leveraging learned visual priors.
To adapt these priors to 3D scenes, existing methods explore camera conditioning~\citep{gao2024cat3d, sargent2024zeronvs}, geometric guidance~\citep{cao2025mvgenmaster, li2025nvcomposer}, cross-view consistency~\citep{kwon2025cameo, muller2024multidiff}, and geometry-aware latent representations~\citep{jang2026repurposing}.
While these approaches primarily focus on view synthesis, Omni-View~\citep{hu2026omni} jointly models novel-view synthesis, 3D language understanding, and geometric estimation, emphasizing how generation facilitates understanding.
Our work instead investigates how semantic and spatial supervision from 3D language understanding can improve novel-view generation.
We pursue this direction through joint training on overlapping scene domains and explicitly use the understanding expert's visual representations during generation.

\subsection{3D Language Understanding}
Understanding 3D scenes through natural language involves diverse tasks such as 3D question answering~\citep{azuma2022scanqa, ma20263d, ma2022sqa3d, yang2025thinking}, 3D dense captioning~\citep{chen2021scan2cap, chen2023end, chen2024vote2cap}, and 3D visual grounding~\citep{chen2020scanrefer, zhang2023multi3drefer}.
A line of open-vocabulary 3D understanding methods lifts semantic knowledge from 2D foundation models, particularly CLIP~\citep{radford2021learning} and LSeg~\citep{li2022language}, into 3D representations such as point clouds~\citep{jatavallabhula2023conceptfusion, yang2024regionplc, zhou2025ov3d}, neural radiance fields~\citep{engelmann2024opennerf, kerr2023lerf}, and 3D Gaussians~\citep{jun2025dr, qin2024langsplat, shi2024language}.
Building on recent developments in 2D Multimodal Large Language Models (MLLMs)~\citep{Qwen2.5-VL, hurst2024gpt, liu2023visual}, numerous efforts aim to extend multimodal understanding to 3D environments by introducing geometric knowledge into these models in various ways.
Some works incorporate spatial information through 3D positional embeddings~\citep{zheng2025video, zhu2024llava}, voxelized scene features~\citep{fu2024scene}, or bird's-eye-view representations~\citep{qi2026gpt4scene}, whereas others inject geometric priors through distillation~\citep{huang20263drs} or fusion~\citep{fan2026vlm, zheng2025learningvideos3dworld} of 3D-aware features.
Ross3D~\citep{wang2025ross3d} and G$^2$VLM~\citep{hu2025g} take a different approach by combining 3D language understanding with reconstruction, exploiting their complementary signals to improve spatial understanding.
Imagine3D-LLM~\citep{jung2026imagine3d} improves 3D language understanding by learning compact Gaussian scene representations through reconstruction supervision.

\subsection{Unified multimodal model}
Unified multimodal models integrate multimodal understanding and visual generation within a single framework.
Chameleon~\citep{team2024chameleon} and Emu3~\citep{wang2024emu3} unify these capabilities through autoregressive next-token prediction.
Janus~\citep{wu2025janus, chen2025janus} uses separate visual encoders for understanding and generation while retaining a unified Transformer backbone.
Other approaches combine next-token prediction with diffusion-based visual generation~\citep{chen2025blip3, xie2025show}.
Bagel~\citep{deng2025emerging} and Lance~\citep{fu2026lance} adopt Mixture-of-Transformers (MoT) architectures with specialized Transformer experts interacting through joint self-attention.
Recent studies also investigate knowledge transfer between understanding and generation~\citep{kang2026transferability}.
Building on this line of research, we focus on improving novel-view generation through semantic and spatial supervision from 3D language understanding.

%% file: sec/appendix/X_C.tex
\newpage
\section{Additional Ablation and Analysis}
\subsection{Ablation on Joint Training Design.}
\input{tables/suppl_joint_nvs}
\input{tables/suppl_joint_qa}

We examine how joint training and visual conditioning affect novel-view generation and 3D language understanding.
\textbf{Joint} denotes training on both tasks. 
Without joint training, the model is trained only on the task being evaluated.
\textbf{Gen-ViT} denotes ViT conditioning for novel-view generation, while \textbf{Und-VAE} denotes additional VAE conditioning for 3D language understanding.
All configurations with Gen-ViT use the same generation-side ViT conditioning, Plücker-ray embeddings, and attention mask. Thus, the Joint \xmark \ and Gen-ViT \cmark \ setting corresponds to Lance-FT + ViT and differs from our final model only in the absence of 3D language understanding training.
We use the same datasets and evaluation protocols as in the main experiments.
Our model enables Joint and Gen-ViT without Und-VAE.

\paragrapht{Novel-view generation.}
Table~\ref{tab:supp_nvs} shows that adding ViT conditioning improves PSNR, SSIM, and LPIPS across all four datasets.
Joint training without Und-VAE further improves all three image fidelity metrics and most geometric consistency metrics compared with generation-only training with ViT conditioning.
However, additionally enabling Und-VAE reduces image fidelity on every dataset and worsens most geometric metrics, despite improvements in rotation RPE on ScanNet and ScanNet++ and translation RPE on ScanNet and ScanNet++.
Our configuration achieves the best image fidelity across all four datasets, supporting the use of joint training with ViT conditioning without additional VAE conditioning for understanding.

\paragrapht{3D language understanding.}
Table~\ref{tab:supp_qa} shows relatively small variations in understanding performance across the evaluated configurations.
Under joint training, additional VAE conditioning yields modest gains on SQA3D and visual grounding, while our configuration performs better on most ScanQA metrics.
Thus, Und-VAE provides no consistent benefit across understanding benchmarks.
Our configuration maintains broadly comparable understanding performance while achieving better novel-view generation quality.

\newpage
\input{figures/tex/supp_corr}

\input{figures/tex/supp_attn}

\subsection{Additional analysis on correspondence of VAE.}

We extend the analysis in Sec.~\ref{sec:analysis} by examining how cross-view correspondence develops during training and visualizing the cosine similarity of VAE token representations across views.
We measure token-level correspondence using the attention-output features at layer $l=26$.

\paragrapht{Correspondence throughout training.}
Figure~\ref{fig:sup_corr} compares \ours with Lance-FT throughout training on ScanNet.
Across all three thresholds, \ours reaches the final correspondence accuracy of Lance-FT substantially earlier and achieves higher accuracy throughout training after the initial stage.
These results suggest that our joint training design promotes the development of cross-view consistent VAE representations.

\paragrapht{Cross-view feature similarity.}
We visualize cross-view feature similarity by computing cosine similarity between a query VAE token's attention-output feature and those of all VAE tokens in another view, without camera conditioning.
The resulting similarity maps are overlaid on the corresponding images.
Figure~\ref{fig:sup_attn} presents the examples of overlaid images of Lance-FT and \ours.
Compared with the diffuse responses of the baseline, \ours exhibits more localized feature responses around corresponding scene regions, illustrating its ability to capture cross-view correspondences more accurately.

\newpage
\input{tables/supp_attn_nvs}
\subsection{Ablation on bidirectional attention between visual tokens.}
\label{supp: attn_mask}
Novel-view generation in \ours involves three groups of visual tokens: source ViT tokens, clean source VAE tokens, and noisy target VAE tokens.
We extend the ablation in Table~\ref{tab:att} by examining the effects of allowing bidirectional attention between different visual token groups.
\textbf{ViT--VAE bidir.} denotes bidirectional attention between source ViT and clean source VAE tokens, while \textbf{Noise--Clean bidir.} denotes bidirectional attention between clean source and noisy target VAE tokens.
Our model enables both forms of bidirectional attention.

Enabling both forms of attention achieves the best results across all reported metrics on DL3DV and RealEstate10K, and most metrics on ScanNet++.
On ScanNet, enabling only Noise--Clean bidirectional attention yields better PSNR, SSIM, and LPIPS, while our configuration achieves better ATE, translation RPE, and MEt3R.
Nevertheless, enabling both connections provides more consistent performance across datasets, including the out-of-domain RealEstate10K and ScanNet++ benchmarks, supporting our attention mask design.

\subsection{Effect of Understanding Supervision.}
\input{tables/supp_data_scale}
\label{supp:data}
We examine how the composition of 3D language understanding supervision affects novel-view generation by comparing question answering, visual grounding, and their combination.
All configurations use the same architecture and ViT conditioning and are evaluated at matched intermediate checkpoints with equal numbers of generation and understanding training samples.

Table~\ref{tab:supp_und_supervision} shows that QA-only supervision achieves the best PSNR, SSIM, and LPIPS on both DL3DV and ScanNet, while grounding-only supervision performs comparably to the combined configuration.
These results suggest that allocating more of the understanding training budget to question answering benefits novel-view generation in this setting.
Our default configuration includes both tasks to support question answering and visual grounding.

\newpage
\subsection{Additional qualitative results.}
\label{supp: qual}
We provide additional qualitative results on DL3DV, ScanNet, Re10K, and ScanNet++ in Figs.~\ref{fig:supp_dl3dv_4}--\ref{fig:supp_2}. 

%% file: tables/suppl_joint_nvs.tex
\begin{table*}[t]
    \caption{
    \textbf{Ablation on joint training settings for novel-view generation.}
        All Gen-ViT configurations use the same generation-side camera conditioning.
    }
    \label{tab:supp_nvs}    
    \centering\small
    \resizebox{\linewidth}{!}{
    \begin{tabular}{c|ccc|ccc|cccc}
    \toprule
    \multirow{2}{*}{Dataset} & \multirow{2}{*}{Joint} & \multirow{2}{*}{Gen-ViT} & \multirow{2}{*}{Und-VAE} & \multicolumn{3}{c|}{\textbf{2D Metrics}} & \multicolumn{4}{c}{\textbf{3D Metrics}} \\
    & & & & PSNR $\uparrow$ & SSIM $\uparrow$ & LPIPS $\downarrow$ & ATE $\downarrow$ & RPE$\mathrm{_r}$ $\downarrow$ & RPE$\mathrm{_t}$ $\downarrow$ & MEt3R $\downarrow$ \\
    \midrule
    \multirow{4}{*}{DL3DV} & \xmark & \xmark & \xmark & 13.291 & 0.297 & 0.432 & \underline{0.185} & \underline{3.80} & \underline{0.354} & \underline{0.361} \\
    & \xmark & \cmark & \xmark & \underline{13.878} & \underline{0.321} & \underline{0.384} & 0.267 & 5.37 & 0.498 & 0.367  \\
    & \cmark & \cmark & \cmark & 13.482 & 0.313 & 0.421 & 0.380 & 5.96 & 0.681 & 0.407  \\
    & \cellcolor{red!10} \cmark & \cellcolor{red!10} \cmark & \cellcolor{red!10} \xmark & \cellcolor{red!10} \textbf{14.939} & \cellcolor{red!10} \textbf{0.355} & \cellcolor{red!10} \textbf{0.313} & \cellcolor{red!10} \textbf{0.158} &\cellcolor{red!10}  \textbf{3.11} & \cellcolor{red!10} \textbf{0.298} & \cellcolor{red!10} \textbf{0.349}\\
    \midrule
    \multirow{4}{*}{ScanNet} & \xmark & \xmark & \xmark & 14.292 & 0.530 & 0.518 & 0.233 & 15.45 & 0.351 & \textbf{0.398}\\
    & \xmark & \cmark & \xmark & \underline{15.574} & \underline{0.560} & \underline{0.438} & \textbf{0.182} & 14.26 & \textbf{0.288} & \underline{0.411}  \\
    & \cmark & \cmark & \cmark & 15.160 & 0.553 & 0.451 & 0.232 & \textbf{12.83} & \underline{0.335} & 0.469  \\
    & \cellcolor{red!10} \cmark & \cellcolor{red!10} \cmark & \cellcolor{red!10} \xmark & \cellcolor{red!10} \textbf{15.747} & \cellcolor{red!10} \textbf{0.564} & \cellcolor{red!10} \textbf{0.434} & \cellcolor{red!10} \underline{0.191} & \cellcolor{red!10} \underline{13.85} & \cellcolor{red!10} 0.340  & \cellcolor{red!10} 0.413 \\
    \midrule
    \multirow{4}{*}{Re10K} & \xmark & \xmark & \xmark & 14.548 & 0.494 & 0.432 & \underline{0.116} & \underline{1.91} & \underline{0.221} & \underline{0.328} \\
    & \xmark & \cmark & \xmark & \underline{15.725} & \underline{0.526} & \underline{0.354} & 0.161 & 2.18 & 0.289 & 0.338  \\
    & \cmark & \cmark & \cmark & 15.082 & 0.519 & 0.392 & 0.182 & 2.68 & 0.335 & 0.363  \\
    & \cellcolor{red!10} \cmark & \cellcolor{red!10}\cmark & \cellcolor{red!10}\xmark &\cellcolor{red!10}\textbf{16.738} & \cellcolor{red!10}\textbf{0.562} & \cellcolor{red!10}\textbf{0.290} & \cellcolor{red!10}\textbf{0.106} & \cellcolor{red!10}\textbf{1.68} & \cellcolor{red!10}\textbf{0.202} & \cellcolor{red!10}\textbf{0.319}\\
    \midrule
    \multirow{4}{*}{ScanNet++} & \xmark & \xmark & \xmark & 14.468 & 0.571 & 0.483 & \underline{0.301} & 19.58 & \underline{0.612} & \underline{0.417}\\
    & \xmark & \cmark & \xmark & \underline{15.030} & \underline{0.582} & \underline{0.451} & 0.308 & 19.09 & 0.655 & 0.426  \\
    & \cmark & \cmark & \cmark & 14.461 & 0.570 & 0.475 & 0.303 & \textbf{16.42} & \textbf{0.589} & 0.475  \\
    & \cellcolor{red!10} \cmark & \cellcolor{red!10}\cmark & \cellcolor{red!10}\xmark & \cellcolor{red!10}\textbf{15.242} & \cellcolor{red!10}\textbf{0.583} & \cellcolor{red!10}\textbf{0.419} & \cellcolor{red!10}\textbf{0.287} & \cellcolor{red!10}\underline{16.54} & \cellcolor{red!10}0.649  & \cellcolor{red!10}\textbf{0.398}\\
    \bottomrule
    \end{tabular}
    }
    \vspace{-5pt}

\end{table*}

%% file: tables/suppl_joint_qa.tex
\begin{table*}[t]
    \caption{
    \textbf{Ablation on joint training settings for 3D language understanding.}
    }
    \label{tab:supp_qa}
    \centering\small
    \resizebox{\textwidth}{!}{
    \begin{tabular}{ccc|ccccccccccc}
    \toprule
    \multirow{2}{*}{Joint} & \multirow{2}{*}{Gen-ViT} & \multirow{2}{*}{Und-VAE} &  \multicolumn{2}{c}{SQA3D$_{\text{test}}$}& \multicolumn{5}{c}{ScanQA$_{\text{val}}$} & \multicolumn{2}{c}{ScanRefer$_{\text{val}}$} & \multicolumn{2}{c}{Multi3DRefer$_{\text{val}}$} \\
    \cmidrule(lr){4-5}
    \cmidrule(lr){6-10}
    \cmidrule(lr){11-12}
    \cmidrule(lr){13-14}
    &&& EM & EM-R & C & B-4 & M & R & EM & Acc@0.25 & Acc@0.5 & F1@0.25 & F1@0.5\\
    \midrule\midrule
    \xmark & \xmark & \xmark & 54.3 & 57.3 & \underline{86.5} & \textbf{13.6} & \underline{17.2} & \underline{42.9} & 24.8 & 52.6 & 47.0 & \underline{53.1} & \underline{48.5} \\
    \xmark & \xmark & \cmark & 54.3 & 57.0 & 85.1 & \textbf{13.6} & 16.9 & 42.3 & \underline{25.0} & 51.9 & 46.3 & 52.8 & 48.3 \\
    \cmark & \cmark & \cmark & \textbf{55.4} & \textbf{58.4} & 84.3 & \underline{11.8} & 16.6 & 42.6 & \textbf{25.6} & \textbf{53.4} & \textbf{47.4} & \textbf{53.2} & \textbf{48.6} \\
    \rowcolor{red!10} \cmark & \cmark & \xmark  & \underline{54.9} & \underline{57.7} & \textbf{87.7} & \textbf{13.6} & \textbf{17.4} & \textbf{43.7} & \textbf{25.6} & \underline{52.9} & \underline{47.3} & 52.6 & 48.3 \\
    \bottomrule
    \end{tabular}
    }
    \vspace{-5pt}

\end{table*}

%% file: figures/tex/supp_corr.tex
\begin{figure*}[t]
    \centering
    \includegraphics[width=\linewidth]{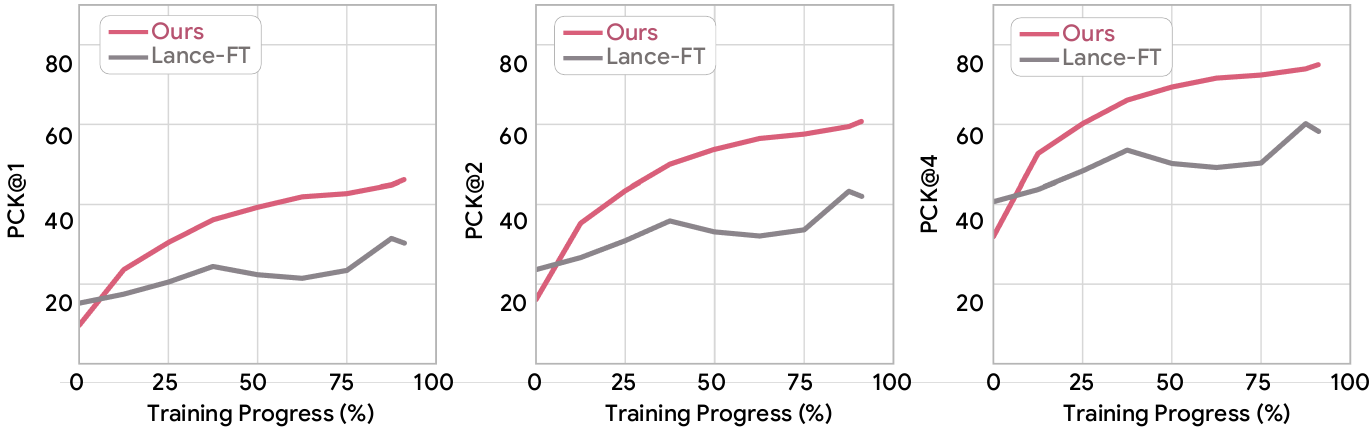}
    \vspace{-10pt}
    \caption{
        \textbf{Cross-view correspondence accuracy throughout training.}
    }
    \label{fig:sup_corr}
\end{figure*}

%% file: figures/tex/supp_attn.tex
\begin{figure*}[t]
    \centering
    \includegraphics[width=\linewidth]{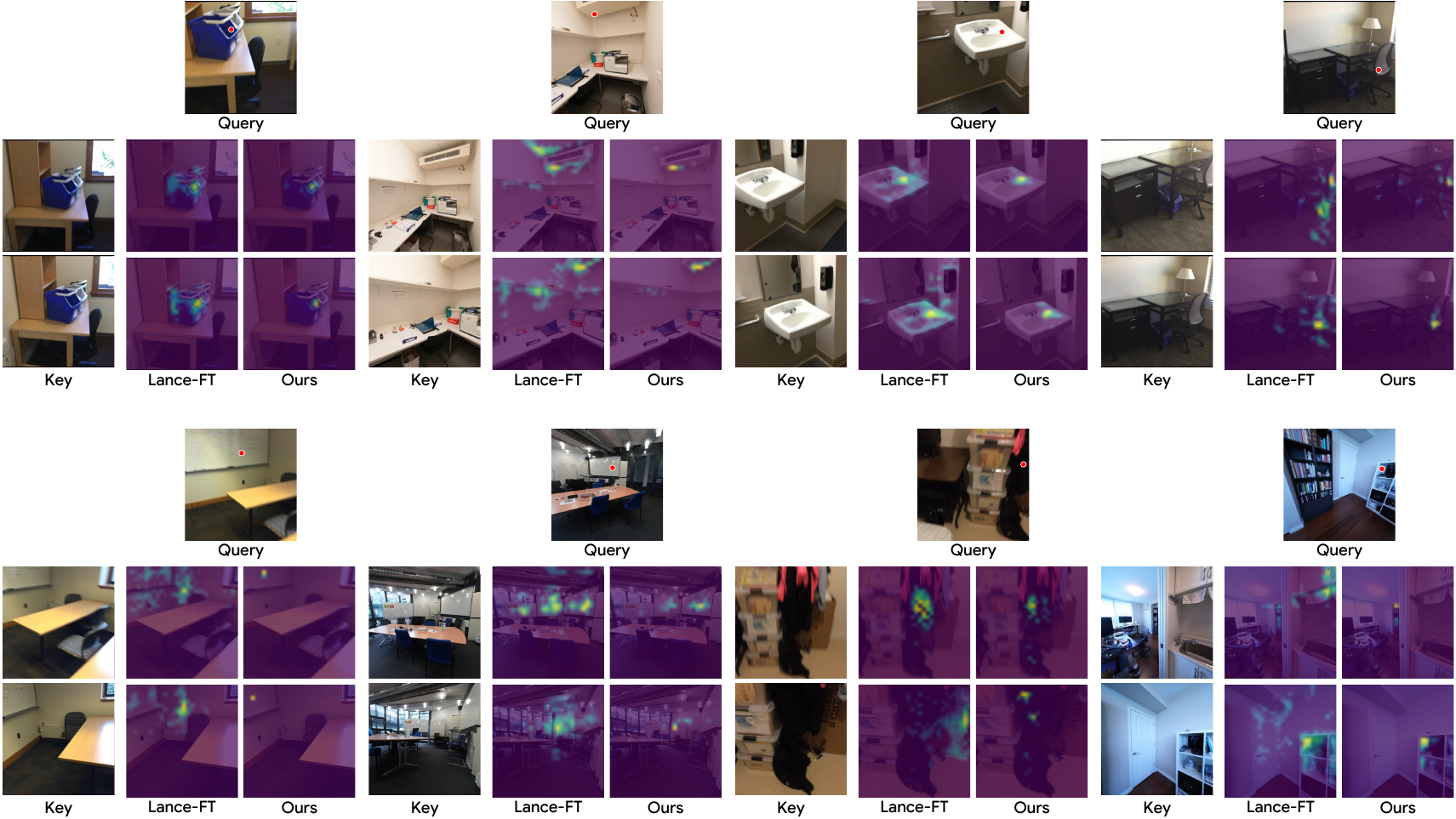}
    \vspace{-10pt}
    \caption{
        \textbf{Visualization of cross-view VAE feature similarity.}
    }
    \label{fig:sup_attn}
    \vspace{-10pt}
\end{figure*}

%% file: tables/supp_attn_nvs.tex
\begin{table*}[t]
    \caption{
    \textbf{Ablation on bidirectional attention between visual tokens.}
    }
    \label{tab:supp_attn}    
    \centering\small
    \resizebox{\linewidth}{!}{
    \begin{tabular}{c|cc|ccc|cccc}
    \toprule
    \multirow{2}{*}{Dataset} & ViT--VAE & Noise--Clean & \multicolumn{3}{c|}{\textbf{2D Metrics}} & \multicolumn{4}{c}{\textbf{3D Metrics}} \\
    & bidir. & bidir. & PSNR $\uparrow$ & SSIM $\uparrow$ & LPIPS $\downarrow$ & ATE $\downarrow$ & RPE$_\mathrm{_r}$ $\downarrow$ & RPE$_\mathrm{_t}$ $\downarrow$ & MEt3R $\downarrow$ \\
    \midrule
    \multirow{4}{*}{DL3DV} & \xmark & \xmark & 13.573 & 0.310 & 0.410 & 0.378 & 6.54 & 0.687 & 0.373  \\
    & \xmark & \cmark & \underline{13.945} & \underline{0.326} & \underline{0.384} & \underline{0.319} & 6.28 &  \underline{0.607} &  \underline{0.363}  \\
    & \cmark & \xmark & 13.653 & 0.314 & 0.424 & 0.409 & \underline{5.59} & 0.748 & 0.406  \\
    & \cellcolor{red!10} \cmark & \cellcolor{red!10} \cmark & \cellcolor{red!10} \textbf{14.939} &\cellcolor{red!10} \textbf{0.355} & \cellcolor{red!10}\textbf{0.313} &\cellcolor{red!10} \textbf{0.158} & \cellcolor{red!10}\textbf{3.11} & \cellcolor{red!10}\textbf{0.298} & \cellcolor{red!10}\textbf{0.349}\\
    \midrule
    \multirow{4}{*}{ScanNet} & \xmark & \xmark & 15.542 & 0.557 & 0.446 & \underline{0.200} & \underline{13.03} & 0.348 & \underline{0.420}  \\
    & \xmark & \cmark & \textbf{16.102} & \textbf{0.575} & \textbf{0.426} & 0.219 & 13.52 & \underline{0.341} & 0.425  \\
    & \cmark & \xmark & 15.488 & \underline{0.564} & 0.446 & 0.221 & \textbf{11.65} & 0.371 & 0.470  \\
    & \cellcolor{red!10} \cmark & \cellcolor{red!10} \cmark  &\cellcolor{red!10}\underline{15.747} & \cellcolor{red!10}\underline{0.564} & \cellcolor{red!10}\underline{0.434} & \cellcolor{red!10}\textbf{0.191} & \cellcolor{red!10}13.85 & \cellcolor{red!10}\textbf{0.340} & \cellcolor{red!10}\textbf{0.413} \\
    \midrule
    \multirow{4}{*}{Re10K} & \xmark & \xmark & 15.549 & 0.527 & 0.367 & 0.179 & \underline{2.16} & 0.321 & \underline{0.346}  \\
    & \xmark & \cmark & \underline{15.988} & \underline{0.544} & \underline{0.347} & \underline{0.168} & 2.40 & \underline{0.315} & \underline{0.346} \\
    & \cmark & \xmark & 15.351 & 0.523 & 0.386 & 0.187 & 2.71 & 0.349 & 0.372  \\
    & \cellcolor{red!10} \cmark & \cellcolor{red!10}\cmark &\cellcolor{red!10}\textbf{16.738} & \cellcolor{red!10}\textbf{0.562} & \cellcolor{red!10}\textbf{0.290} & \cellcolor{red!10}\textbf{0.106} & \cellcolor{red!10}\textbf{1.68} & \cellcolor{red!10}\textbf{0.202} & \cellcolor{red!10}\textbf{0.319}\\
    \midrule
    \multirow{4}{*}{ScanNet++} & \xmark & \xmark & 14.848 & 0.582 & 0.465 & 0.330 & \underline{17.55} & \textbf{0.640} & 0.446  \\
    & \xmark & \cmark & \underline{15.178} & \textbf{0.594} & \underline{0.447} & \underline{0.317} & 22.32 & 0.685 & \underline{0.429}  \\
    & \cmark & \xmark & 14.703 & 0.580 & 0.479 & 0.332 & 20.28 & 0.672 & 0.488  \\
    & \cellcolor{red!10} \cmark & \cellcolor{red!10}\cmark & \cellcolor{red!10}\textbf{15.242} & \cellcolor{red!10}\underline{0.583} & \cellcolor{red!10}\textbf{0.419} & \cellcolor{red!10}\textbf{0.287} & \cellcolor{red!10}\textbf{16.54} & \cellcolor{red!10}\underline{0.649}& \cellcolor{red!10}\textbf{0.398}\\
    \bottomrule
    \end{tabular}
    }
    \vspace{-5pt}

\end{table*}

%% file: tables/supp_data_scale.tex
\begin{wraptable}{r}{0.5\textwidth}
    \vspace{-10pt}
    \centering

    \caption{
        \textbf{Ablation on the composition of 3D language understanding supervision.}
    }
    \label{tab:supp_und_supervision}

    \vspace{-3pt}
    \small

    \resizebox{\linewidth}{!}{
    \begin{tabular}{c|cc|ccc}
        \toprule
        Dataset & QA & Grounding
        & PSNR $\uparrow$ & SSIM $\uparrow$ & LPIPS $\downarrow$ \\
        \midrule
        \multirow{3}{*}{DL3DV}
        & \cmark & \xmark
        & \textbf{13.66} & \textbf{0.313} & \textbf{0.402} \\
        & \xmark & \cmark
        & \underline{13.03} & \underline{0.291} & 0.453 \\
        & \cmark & \cmark
        & 12.92 & 0.285 & \underline{0.451} \\
        \midrule
        \multirow{3}{*}{ScanNet}
        & \cmark & \xmark
        & \textbf{13.94} & \textbf{0.525} & \textbf{0.524} \\
        & \xmark & \cmark
        & 13.50 & \underline{0.514} & 0.552 \\
        & \cmark & \cmark
        & \underline{13.54} & 0.510 & \underline{0.546} \\
        \bottomrule
    \end{tabular}
    }

    \vspace{-8pt}
\end{wraptable}

%% file: figures/tex/supp_qual_dl3dv_s4.tex
\begin{figure*}[t]
    \centering
    \includegraphics[width=\linewidth]{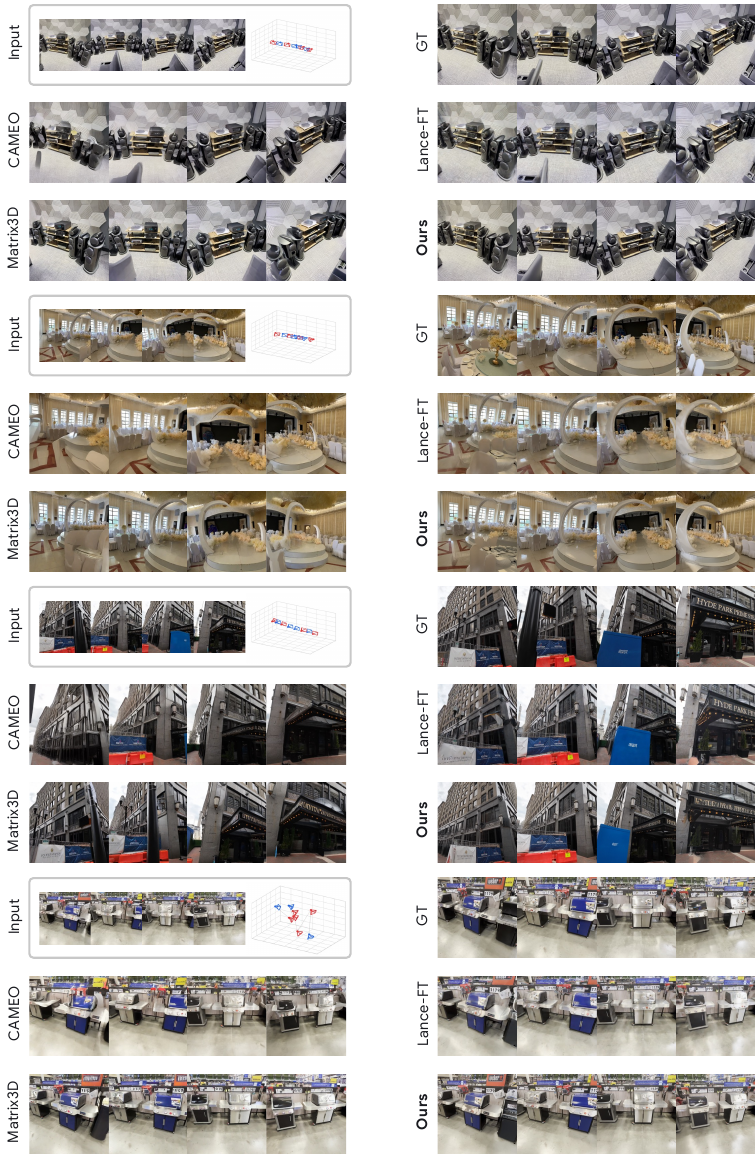}
    \vspace{-10pt}
    \caption{
        \textbf{Additional qualitative results of novel view generation on DL3DV,} comparing \ours with baseline methods. \textcolor{blue}{Blue} and \textcolor{red}{red} cameras indicate source and target viewpoints, respectively.
    }
    \label{fig:supp_dl3dv_4}
    \vspace{-10pt}
\end{figure*}

%% file: figures/tex/supp_qual_scannet_s4.tex
\begin{figure*}[t]
    \centering
    \includegraphics[width=\linewidth]{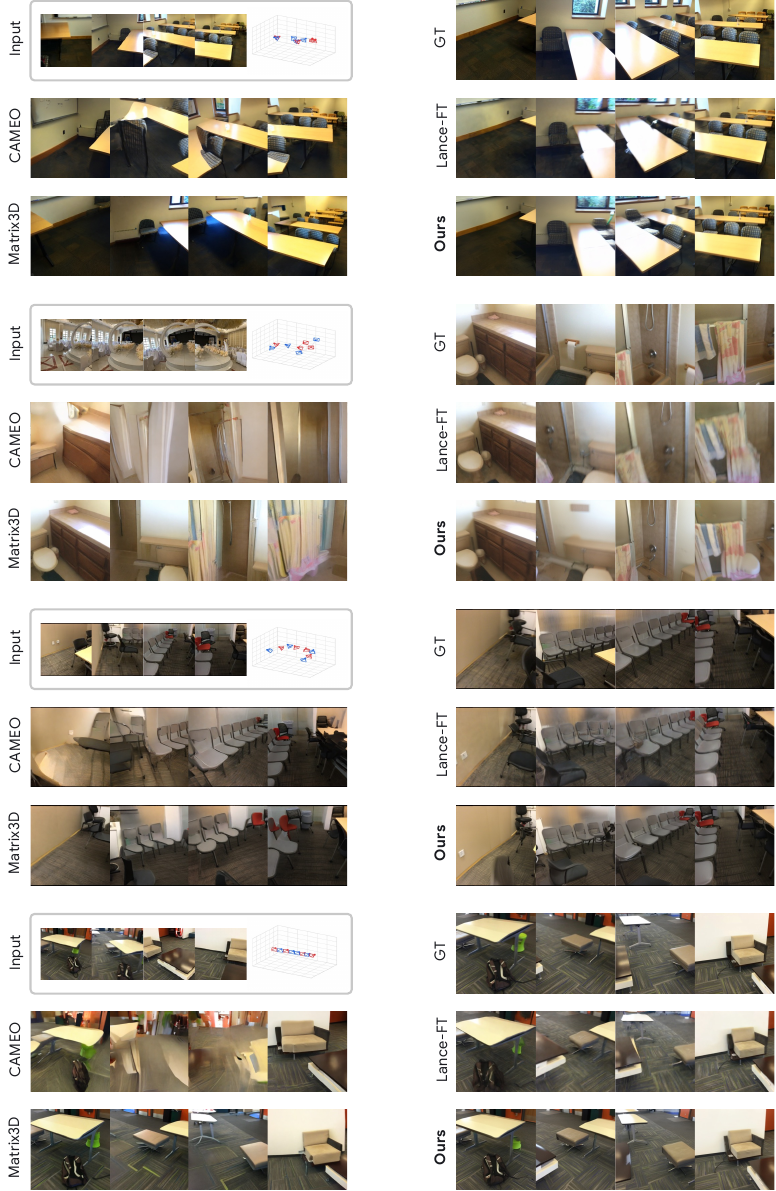}
    \vspace{-10pt}
    \caption{
        \textbf{Additional qualitative results of novel view generation on ScanNet,} comparing \ours with baseline methods. \textcolor{blue}{Blue} and \textcolor{red}{red} cameras indicate source and target viewpoints, respectively.
    }
    \label{fig:supp_scannet_s4}
    \vspace{-10pt}
\end{figure*}

%% file: figures/tex/supp_qual_re10k_s4.tex
\begin{figure*}[t]
    \centering
    \includegraphics[width=\linewidth]{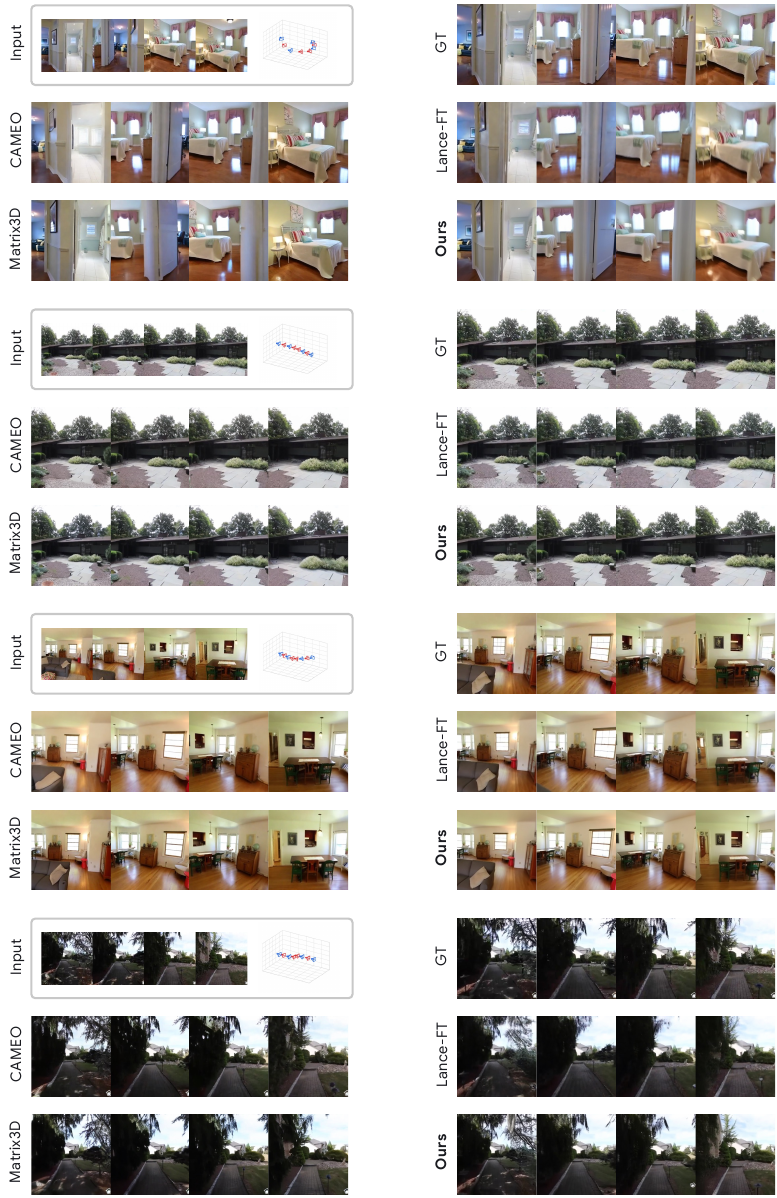}
    \vspace{-10pt}
    \caption{
        \textbf{Additional qualitative results of novel view generation on Re10K,} comparing \ours with baseline methods. \textcolor{blue}{Blue} and \textcolor{red}{red} cameras indicate source and target viewpoints, respectively.
    }
    \label{fig:supp_re10k_4}
    \vspace{-10pt}
\end{figure*}

%% file: figures/tex/supp_qual_pp_4.tex
\begin{figure*}[t]
    \centering
    \includegraphics[width=\linewidth]{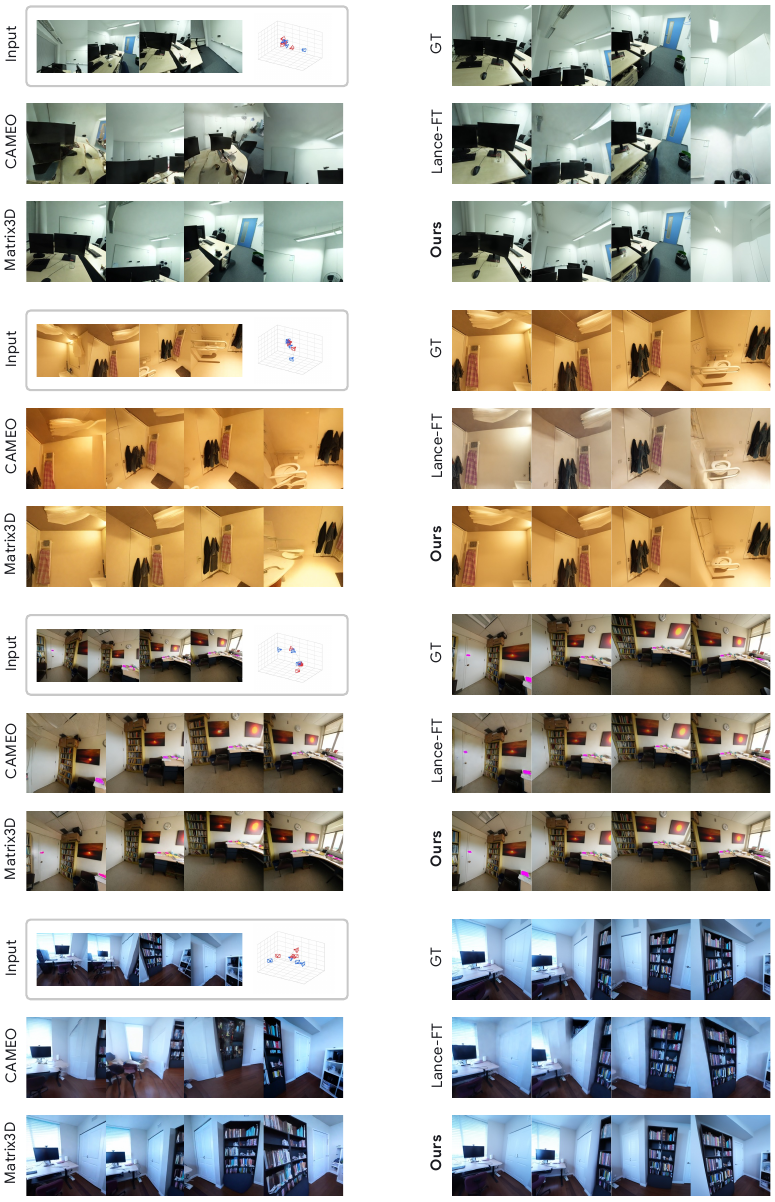}
    \vspace{-10pt}
    \caption{
        \textbf{Additional qualitative results of novel view generation on ScanNet++,} comparing \ours with baseline methods. \textcolor{blue}{Blue} and \textcolor{red}{red} cameras indicate source and target viewpoints, respectively.
    }
    \label{fig:supp_scannetpp_s4}
    \vspace{-10pt}
\end{figure*}

%% file: figures/tex/supp_source_3.tex
\begin{figure*}[t]
    \centering
    \includegraphics[width=\linewidth]{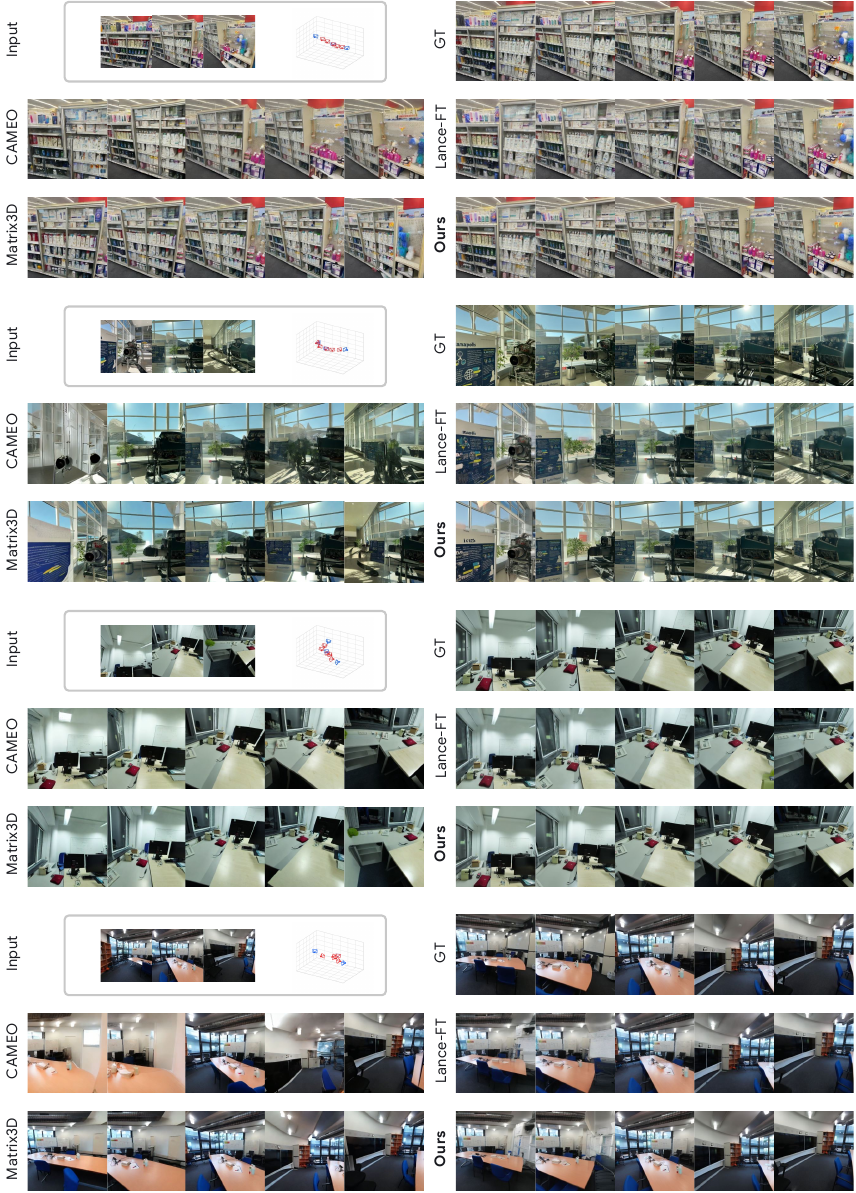}
    \vspace{-10pt}
    \caption{
        \textbf{Additional qualitative results of novel view generation on $N=3$ source views,} comparing \ours with baseline methods. \textcolor{blue}{Blue} and \textcolor{red}{red} cameras indicate source and target viewpoints, respectively.
    }
    \label{fig:supp_3}
    \vspace{-10pt}
\end{figure*}

%% file: figures/tex/supp_source_2.tex
\begin{figure*}[t]
    \centering
    \includegraphics[width=\linewidth]{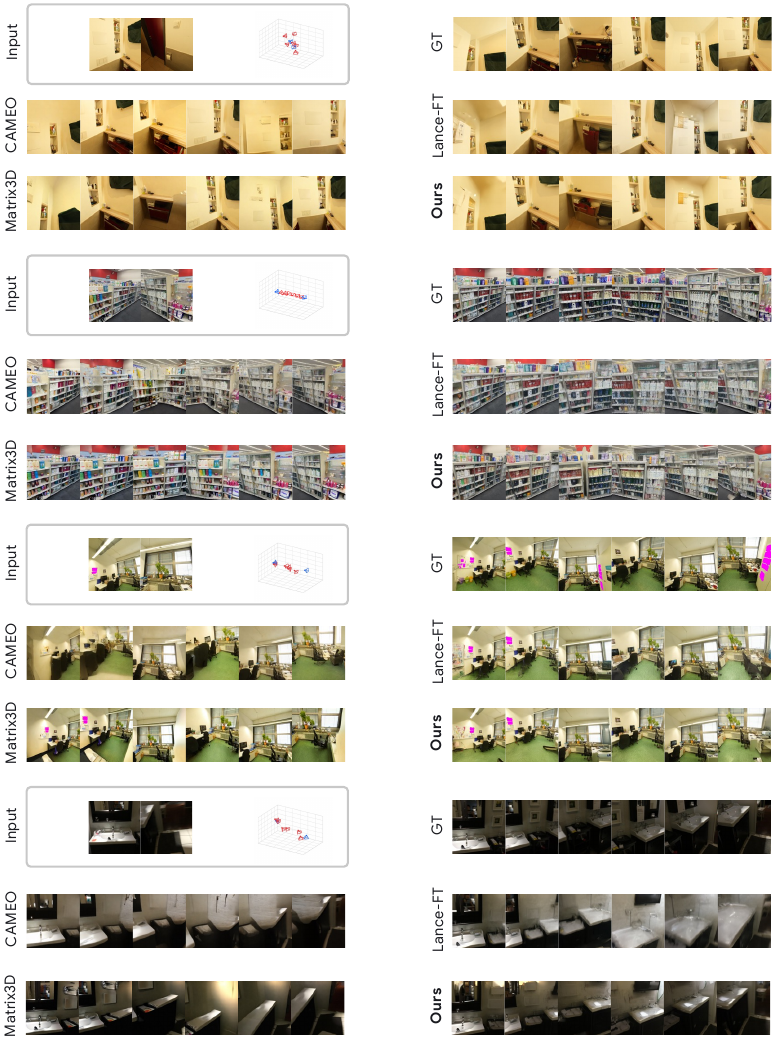}
    \vspace{-10pt}
    \caption{
        \textbf{Additional qualitative results of novel view generation on $N=2$ source views,} comparing \ours with baseline methods. \textcolor{blue}{Blue} and \textcolor{red}{red} cameras indicate source and target viewpoints, respectively.
    }
    \label{fig:supp_2}
    \vspace{-10pt}
\end{figure*}